\documentclass[11pt,a4paper]{article}
\usepackage[hyperref]{emnlp2020}
\usepackage{times}
\usepackage{latexsym}
\usepackage{amssymb}
\usepackage{tabularx}
\usepackage{mathtools}
\usepackage{booktabs}
\usepackage{url}
\usepackage{longtable}
\usepackage{tabu}
\usepackage{multirow}
\usepackage{amsfonts}
\usepackage{tabu}
\usepackage{algorithm}
\usepackage{bbm}

\usepackage{microtype}
\usepackage{amsmath}
\usepackage{subfigure}
\usepackage{enumitem}
\usepackage{adjustbox}
\usepackage{transparent}

\usepackage{tikz}
\usetikzlibrary{shapes.geometric}
\usetikzlibrary{patterns}
\usepackage{pgfplots}
\usetikzlibrary{backgrounds}
\usetikzlibrary{matrix,calc}
\usepackage{bbm}
\usepackage{pgfplots}

\tikzset{fit to page/.style={fit only width,fit only height},
         fit only width/.style={
                    trim left={($(current bounding box.center)-(0.5*\columnwidth,0)$)},
                    trim right={($(current bounding box.center)+(0.5*\columnwidth,0)$)},
          },
          fit only height/.style={
                     execute at end picture={%
                     \useasboundingbox let 
                           \p1=(current bounding box.north east),
                           \p2=(current bounding box.south west) in
                           \pgfextra{\pgfresetboundingbox}
                           (\x2,-0.5*\textheight) rectangle (\x1-(0,0.5*\textheight);
                    }
          }
}

\usepackage{tipa}

\usepackage[T2A,LGR,T1]{fontenc}
\usepackage[utf8]{inputenc}
\usepackage[russian,greek,english]{babel}

\usepackage[normalem]{ulem}

\newcommand\blfootnote[1]{%
  \begingroup
  \renewcommand\thefootnote{}\footnote{\noindent #1}%
  \addtocounter{footnote}{-1}%
  \endgroup
}

\aclfinalcopy 
\def\aclpaperid{2064} 

\title{Automatic Extraction of Rules Governing Morphological Agreement}

\author{Aditi Chaudhary\textsuperscript{1},
    Antonios Anastasopoulos\textsuperscript{2,$\dagger$},
  Adithya Pratapa\textsuperscript{1},
  David R. Mortensen\textsuperscript{1}, \\
  {\bf Zaid Sheikh\textsuperscript{1}, Yulia Tsvetkov\textsuperscript{1}, Graham Neubig\textsuperscript{1}}  \\
  \textsuperscript{1}Language Technologies Institute, Carnegie Mellon University\\
  \textsuperscript{2}Department of Computer Science, George Mason University\\
  {\footnotesize \texttt{\{aschaudh,vpratapa,dmortens,zsheikh,ytsvetko,gneubig\}@cs.cmu.edu}} \hspace{.5cm}
  {\footnotesize \texttt{antonis@gmu.edu}}
 }
\date{}

\begin{document}
\maketitle
\begin{abstract}
Creating a descriptive grammar of a language is an indispensable step for language documentation and preservation. However, at the same time it is a tedious, time-consuming task. In this paper, we take steps towards automating this process by devising an automated framework for extracting a first-pass grammatical specification from raw text in a concise, human- \textit{and} machine-readable format. We focus on extracting rules describing \textit{agreement}, a morphosyntactic phenomenon at the core of the grammars of many of the world's languages.
We apply our framework to all languages included in the Universal Dependencies project, with promising results.
Using cross-lingual transfer, even with no expert annotations in the language of interest, our framework extracts a grammatical specification which is nearly equivalent to those created with large amounts of gold-standard annotated data.
We confirm this finding with human expert evaluations of the rules that our framework produces, which have an average accuracy of~78\%. We release an interface demonstrating the extracted rules at \url{https://neulab.github.io/lase/}. The code is publicly available here.\footnote{\url{https://github.com/Aditi138/LASE-Agreement}}\blfootnote{$\dagger$: Work done at Carnegie Mellon University.}

\end{abstract}
\section{Introduction}


While the languages of the world are amazingly diverse, one thing they share in common is their adherence to grammars --- sets of morpho-syntactic rules specifying how to create sentences in the language.
Hence, an important step in the understanding and documentation of languages is the creation of a \textit{grammar sketch}, 
a concise and human-readable description of the unique characteristics of that particular language (e.g. \citet{huddleston2002cambridge} for English, or \citet{brown2010concise} for the world's languages).

One aspect of morphosyntax that is widely described in such grammatical specifications is \emph{agreement}, the process wherein a word or morpheme selects morphemes in correspondence with another word or phrase in the sentence \cite{corbett2009agreement}. 
Languages have varying degrees of agreement ranging from none (e.g.~Japanese, Malay) to a large amount (e.g.~Hindi, Russian, Chichewa). Patterns of agreement also vary across syntactic subcategories. For instance, regular verbs in English agree with their subject in number and person but modal verbs such as ``will'' show no agreement.

Having a concise description of these rules is of obvious use not only to linguists but also language teachers and learners. Furthermore, having such descriptions in machine-readable format will further enable applications in natural language processing (NLP) such as identifying and mitigating gender stereotypes in morphologically rich languages \cite{zmigrod2019counterfactual}.

\begin{figure*}
\small
\includegraphics[width=\textwidth]{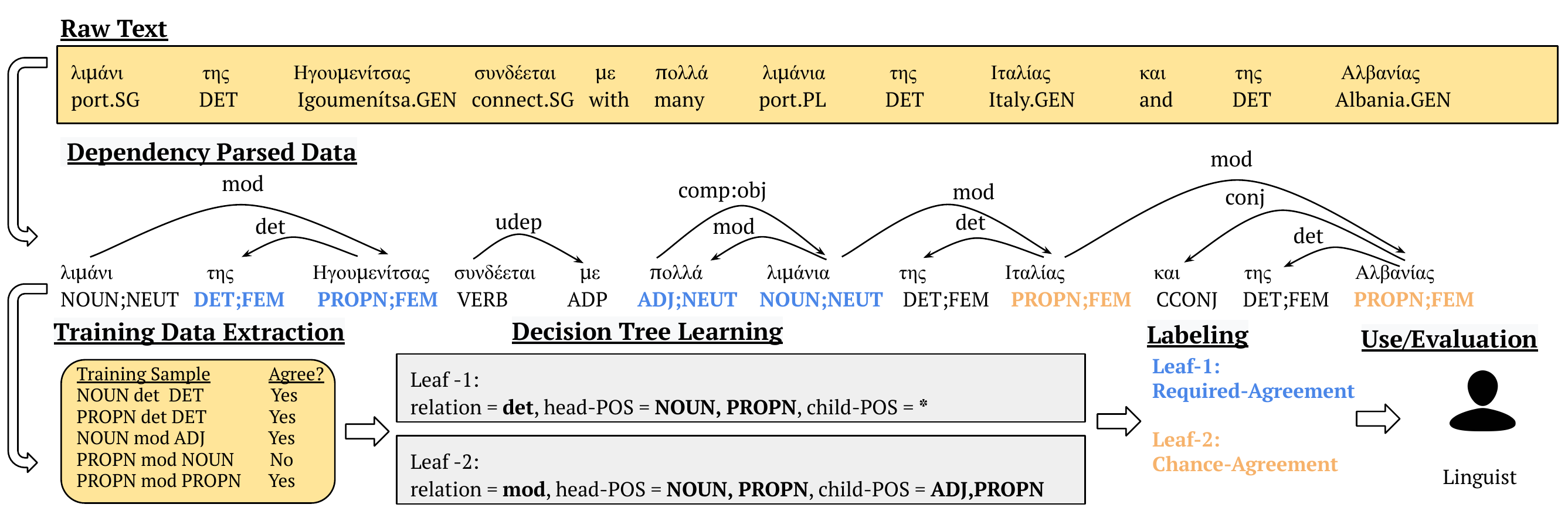}
\caption{An overview of our method's workflow for gender agreement in Greek. The example sentence translates to ``The port of Igoumenitsa is connected to many ports in Italy and Albania.'' First, we dependency parse and morphologically analyze raw text to create training data for our binary agreement classification task. Next, we learn a decision tree to extract the rule set governing gender agreement, and label the extracted leaves as either representing required or chance agreement. Finally these rules are presented to a linguist for perusal.}
\label{fig:agreement}
\end{figure*}

The notion of describing a language ``in its own terms'' based solely on raw data has an established tradition in descriptive linguistics (e.g. \citet{harris1951methods}).
In this work we present a framework (outlined in Figure \ref{fig:agreement}) that \emph{automatically} creates a first-pass specification of morphological agreement rules for various morphological features (Gender, Number, Person, etc.) from a raw text corpus for the language in question.  First, we perform syntactic analysis, predicting part-of-speech (POS) tags, morphological features, and dependency trees. Using this analyzed data, we then learn an agreement prediction model that contains the desired rules. Specifically, we devise a binary classification problem of identifying whether agreement will be observed between a head and its dependent token on a given morphological property. We use decision trees as our classification model because they are easy to interpret and we can easily \emph{extract} the classification rules from the tree leaves to get an initial set of potential agreement rules. Finally, we perform \emph{rule labeling} of the extracted rules, identifying which tree leaves correspond to probable agreement.
This is required because not all agreeing head/dependent token pairs are necessarily due to some underlying rule. For instance, in Figure~\ref{fig:agreement}'s example of Greek gender agreement, both the head and its dependent token \textit{\foreignlanguage{greek}{Ιταλίας}}$\rightarrow$\textit{\foreignlanguage{greek}{Αλβανίας}} have feminine gender, but this agreement is purely bychance, as correctly identified by our framework.

The quality of the learnt rules depends crucially on the quality and quantity of dependency parsed data, which is often not readily available for low-resource languages. Therefore, we experiment with not only gold-standard treebanks, but also trees generated automatically using models trained using cross-lingual transfer learning. This assesses the applicability of the proposed method in a situation where a linguist may want to explore the characteristics of agreement in a language that does not have a large annotated dependency treebank.

We evaluate the correctness of the extracted rules conducting human evaluation with linguists for Greek, Russian, and Catalan. In addition to the manual verification, we also devise a new metric for automatic evaluation of the rules over unseen test data.
Our contributions can be summarized to:
\begin{enumerate}[noitemsep,nolistsep,leftmargin=*]
    \item We propose a framework to automatically extract agreement rules from raw text, and release these rules for 55 languages as part of an interface\footnote{\url{https://neulab.github.io/lase/}} which visualizes the rules in detail along with examples and counter-examples.
    \item We design a human evaluation interface to allow linguists to easily verify the extracted rules. Our framework produces a decent first-pass grammatical specification with the extracted rules having an average accuracy of 78\%. 
    We also devise an automated metric to evaluate our framework when 
    human evaluation is infeasible.
    \item We evaluate the quality of extracted rules under real zero-shot conditions (on Breton, Buryat, Faroese, Tagalog, and Welsh) as well as low-resource conditions (with simulation experiments on Spanish, Greek, Belarusian and Lithuanian) varying the amount of training data. Using cross-lingual transfer, rules extracted with as few as~50 sentences with gold-standard syntactic analysis are nearly equivalent to the rules extracted when we have hundreds/thousands of gold-standard data available. 
\end{enumerate}

\section{Problem Formulation}
\label{model}

For a head $h$ and a dependent $d$ that are in a dependency relation $r$, we will say that they \emph{agree} on a morphological property $f$ if they share the same value for that particular property i.e. $f_h = f_d$.
Some agreements that we observe in parsed data can be attributed to an underlying grammatical rule.
For example, in Figure \ref{fig:agreementexample} the Spanish A.1 shows an example of where subject (\textit{enigmas)} and verb (\textit{son}) need to agree on number. We will refer to such rules as \textit{required-agreement}. Such a required agreement rule dictates that an example like A.2 is ungrammatical and would not appear in well-formed Spanish sentences, since the subject and the verb do not have the same number marking.
However, not all word pairs that agree do so because of some underlying rule, and we will refer to such cases as \textit{chance-agreement}. For example, in Figure \ref{fig:agreementexample} the object (\textit{perro}) and verb (\textit{tiene}) in B.1 only agree in number by chance, and example B.2 (where the object of a singular verb is plural) is perfectly acceptable.

Our goal is to extract, from textual examples, the set of \emph{rules} $\mathcal{R}_l^f$ that concisely describe the agreement process for language $l$.
Concretely, this will indicate for which head-dependent pairs the language displays \textit{required-agreement} and for which we will observe at most \textit{chance-agreement}.
Canonically, agreement rules are defined over syntactic features of a language as seen in Figure \ref{fig:agreementexample} where we have the following rule for Spanish: ``subjects agree with their verbs on number''.\footnote{Sometimes semantic features are used for agreement for eg. \emph{United Nations \underline{is}}, despite \emph{United Nations} being plural, it is treated as singular for purposes of agreement.}
To formalize this notion, we define a \emph{rule} to be a set of features which are defined over the dependency relation, head and dependent token types.
In this paper, we make the simplifying assumption that head and dependent tokens are represented by only part-of-speech features, as we would like our extracted rules to be \emph{concise} and easily interpretable downstream, although this assumption could be relaxed in future work. 

\begin{figure}[t]
    \small
    \begin{tabular}{@{}lllll}
        \multirow{1}{*}{A.1} & Los & enigmas & son & f\'{a}ciles \\
         & \small{DET.PL} & \small riddle.PL & \small be.PL & \small easy.PL \\
         & & \multicolumn{2}{l}{\textcolor{blue}{ \hspace{0.9cm} \rotatebox[origin=c]{90}{$\Rsh$}\raisebox{.25em}{\_}req\raisebox{.25em}{\_\_}\rotatebox[origin=c]{270}{$\Lsh$}}}\\ 
         & \multicolumn{4}{l}{`The riddles are easy.'}\\
         \midrule
        \multirow{1}{*}{A.2} & *Los & enigmas & es & f\'{a}cil \\
         & \small{DET.PL} \ \  & \small riddle.PL & \small be.SG & \small easy.SG \\
         & & \multicolumn{2}{l}{\textcolor{black}{\hspace{0.95cm} \rotatebox[origin=c]{90}{$\Rsh$}wrong\rotatebox[origin=c]{270}{$\Lsh$}}}\\ 
        \midrule
    \end{tabular}
    \begin{tabular}{@{}llllll@{}}
        \small B.1 & Mi & hermano & tiene & un & perro \\
        & \small{My} & \small brother.SG  & \small has.SG  & \small{ART.SG}  & dog.SG \\ 
        & &  \multicolumn{4}{l}{\textcolor{orange}{\textcolor{blue}{\hspace{1.1cm} \rotatebox[origin=c]{90}{$\Rsh$}\raisebox{.25em}{\_}req\raisebox{.25em}{\_\_}\rotatebox[origin=c]{270}{$\Lsh$}}  \rotatebox[origin=c]{90}{$\Rsh$}\raisebox{.25em}{\_\_\_\_\_}chance\raisebox{0.25em}{\_\_\_\_\_}\rotatebox[origin=c]{270}{$\Lsh$}}}\\ 
        & \multicolumn{5}{l}{`My brother has a dog.'}\\
        \midrule
        \small B.2 & Mi & hermano & tiene & muchos & perros \\
        & \small{My} & \small brother.SG  & \small has.SG  & \small{many.PL}  & dog.PL \\ 
        & &  \multicolumn{4}{l}{\textcolor{black}{\textcolor{blue}{\hspace{1.1cm} \rotatebox[origin=c]{90}{$\Rsh$}\raisebox{.25em}{\_}req\raisebox{.25em}{\_\_}\rotatebox[origin=c]{270}{$\Lsh$}}  \rotatebox[origin=c]{90}{$\Rsh$}\raisebox{.25em}{\_\_\_\_\_}correct\raisebox{.25em}{\_\_\_\_\_}\rotatebox[origin=c]{270}{$\Lsh$}}}\\ 
        & \multicolumn{5}{l}{`My brother has many dogs.'}\\
    \end{tabular}
    \caption{Subject-verb number agreement is required in Spanish, as in example A.1, which renders example A.2 ungrammatical. Object-verb agreement is not required, so both B.1 and B.2 are grammatical. The object and the verb in B.1 only agree by chance. 
    }
    \label{fig:agreementexample}
    \vspace{-1em}
\end{figure}

The rule discovery process consists of two major steps: a \emph{rule extraction} step followed by a \emph{rule labeling} and \textit{merging} step (also see Figure~\ref{fig:agreement}).

\subsection{Rule Extraction}
To create our training data for rule extraction, we first annotate raw text with part-of-speech (POS) tags, morphological analyses, and dependency trees.
We then base our training data on these annotations by converting each dependency relation into a triple $\langle{h}, {d}, {r} \rangle$, indicating the head token,  dependent/child token, and dependency relation between ${h}$ and ${d}$ respectively.
From the whole treebank, we now have input features $X_f=\{\langle{h_1}, {d_1}, {r_1} \rangle, \ldots, \langle{h_n}, {d_n}, {r_n}\rangle \}$ and binary output labels $Y\!=\!y_1, \ldots, y_n{}$, where if the head and the dependent token agree on feature $f$ (such that $f_h\!=\!f_d$) we set $y=1$, otherwise $y=0$. We filter out the tuples where either of the linked tokens does not display the morphological feature $f$.

\begin{figure*}
\newcommand{\leaf}[7]{
\node at (0,0) [rectangle,draw,align=left,fill=#7] (leaf#1) {\small Leaf #1: \textbf{not agree: #2, agree: #3}\\[.5em]
\begin{tabular}{@{}r@{\,=\,}p{3cm}}
relation & #4\\
head-pos & #5\\
child-pos & #6
\end{tabular}
};
}
\newcommand{\shortleaf}[7]{
\node at (0,0) [rectangle,draw,align=left,fill=#7] (leaf#1) {\small Leaf #1: \textbf{not agree: #2, agree: #3}\\[.5em]
\begin{tabular}{@{}r@{\,=\,}l@{}}
relation & #4\\
head-pos & #5\\
child-pos & #6
\end{tabular}
};
}
\newcommand{\lleaf}[6]{
\node at (0,0) [rectangle,draw=#6,align=left,fill=#6!20] (leaf#1) {\small Leaf #1: \textbf{#2}\\[.5em]
\begin{tabular}{@{}r@{\,=\,}p{2.5cm}}
relation & #3\\
head-pos & #4\\
child-pos & #5
\end{tabular}
};
}
\newcommand{\shortlleaf}[6]{
\node at (0,0) [rectangle,draw=#6,align=left,fill=#6!20] (leaf#1) {\small Leaf #1: \textbf{#2}\\[.5em]
\begin{tabular}{@{}r@{\,=\,}l@{}}
relation & #3\\
head-pos & #4\\
child-pos & #5
\end{tabular}
};
}

\begin{center}
\resizebox{\linewidth}{!}{
\begin{tabular}{@{}cc@{}c@{}}
\begin{tikzpicture}
    \node at (-4,5) [rectangle,draw,fill=gray!20] (node1) {node 1};
    \node at (-4,2) [rectangle,draw,fill=gray!20] (node2) {node 2};
    \begin{scope}[xshift=-.2cm,yshift=3cm]
    \leaf{3}{778}{58076}{\textit{any}}{\textit{any}}{aux,adj,verb,pron, propn,det,num}{Gray!20};
    \end{scope}
    \begin{scope}[xshift=-6cm,yshift=0cm]
    \shortleaf{1}{1462}{2433}{conj, det}{\textit{any}}{noun}{Gray!20};
    \end{scope}
    \begin{scope}[xshift=0cm,yshift=0cm]
    \shortleaf{2}{268}{373}{comp:obj}{\textit{any}}{noun}{Gray!20};
    \end{scope}
    \draw[thick,->] (node1.south) -- (node2.north) node[midway,left] {\small child-pos= noun};
    \draw[thick,->] (node1.south) -- (leaf3.north west) node[midway,right,text width=5cm,text ragged,yshift=.3cm] {\small child-pos= aux,adj,verb,pron,\\[-.5em] \hspace{1.5cm} propn,det,num};
    \draw[thick,->] (node2.south) -- (leaf1.north) node[midway,left,xshift=-0.3cm] {\small relation= det};
    \draw[thick,->] (node2.south) -- (leaf2.north west) node[midway,right,xshift=0.2cm] {\small relation= comp:obj};
\end{tikzpicture}
&
\begin{tikzpicture}
    \node at (-4,5) [rectangle,draw,fill=gray!20] (node1) {node 1};
    \node at (-4,2) [rectangle,draw,fill=gray!20] (node2) {node 2};
    \begin{scope}[xshift=0cm,yshift=3cm]
    \lleaf{3}{required-agreement}{\textit{any}}{\textit{any}}{aux,adj,verb,pron, propn,det,num}{Blue};
    \end{scope}
    \begin{scope}[xshift=-4cm,yshift=0cm]
    \shortlleaf{1}{chance-agreement}{conj, det}{\textit{any}}{noun}{Dandelion};
    \end{scope}
    \begin{scope}[xshift=0cm,yshift=0cm]
    \shortlleaf{2}{chance-agreement}{comp:obj}{\textit{any}}{noun}{Dandelion};
    \end{scope}
    \draw[thick,->] (node1.south) -- (node2.north) node[midway,right,align=left,text width=2cm] {\small child-pos= noun};
    \draw[thick,->] (node1.south) -- (leaf3.north west) node[midway,right,text width=5cm,text ragged,yshift=.3cm] {\small child-pos= aux,adj,verb,pron,\\[-.5em] \hspace{1.5cm} propn,det,num};
    \draw[thick,->] (node2.south) -- (leaf1.north) node[midway,left,xshift=-0cm] {\small relation= det};
    \draw[thick,->] (node2.south) -- (leaf2.north west) node[midway,right,xshift=0.2cm] {\small relation= comp:obj};
\end{tikzpicture}
&
\begin{tikzpicture}
    \node at (-4,5) [rectangle,draw,fill=gray!20] (node1) {node 1};
    \begin{scope}[xshift=0cm,yshift=3cm]
    \lleaf{3}{required-agreement}{\textit{any}}{\textit{any}}{aux,adj,verb,pron, propn,det,num}{Blue};
    \end{scope}
    \begin{scope}[xshift=-1.5cm,yshift=0cm]
    \shortlleaf{1}{chance-agreement}{conj, det, comp:obj}{\textit{any}}{noun}{Dandelion};
    \end{scope}
    \draw[thick,->] (node1.south) -- (leaf3.north west) node[midway,right,text width=5cm,text ragged,yshift=.3cm] {\small child-pos= aux,adj,verb,pron,\\[-.5em] \hspace{1.5cm} propn,det,num};
    \draw[thick,->] (node1.south) -- (leaf1.north west) node[midway,right,align=left,text width=2cm] {\small child-pos= noun};
\end{tikzpicture}\\
\large{(a) Rule Extraction} & \large{(b) Rule Labeling} & \large{(c) Rule Merging}\\
\end{tabular}%
}
\vspace{-.5cm}
\end{center}

    \caption{Extracting gender agreement rules in Spanish. (a) A decision tree is learned over dependency link triples, inducing a distribution of agreement over examples in each leaf. However, simple majority voting leads to false positives: Leaf-1 includes more agreeing data points, but in reality this agreement is purely by chance. (b) With a statistically-inspired threshold to label the leaves, Leaf-1 gets correctly labeled as \textit{chance-agreement}. (c) We merge leaves with the same label to get a concise representation. Every dependency link triple receives the label of the unique leaf it falls under.}
    \label{fig:tree}
    \vspace{-1em}
\end{figure*}
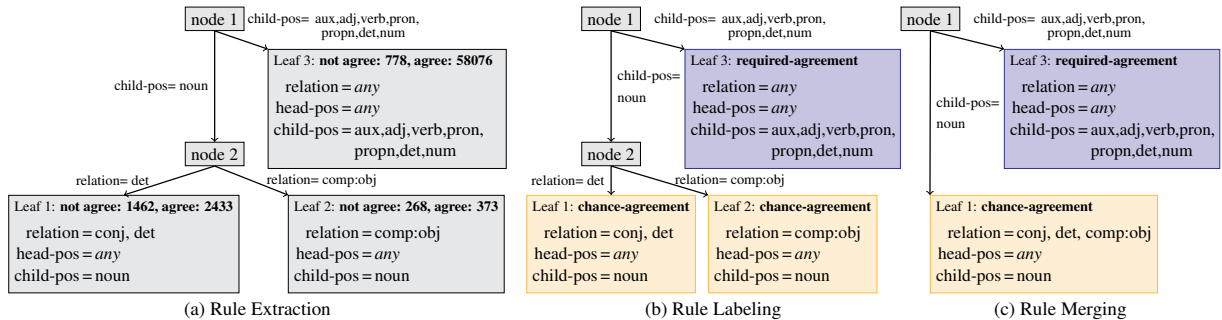

We train a model for $p(Y|X)$ using decision trees \cite{quinlan1986induction} using the CART algorithm \cite{breiman1984classification}. 
A major advantage of decision trees is that they are easy to interpret and we can visualize the exact features used by the decision tree to split nodes.
The decision tree induces a distribution  of agreement over training samples in each leaf, e.g. 99\% agree, 1\% not agree in Leaf-3 for gender agreement in Spanish (Figure \ref{fig:tree}(a)).

\subsection{Rule Labeling}
\label{filtration}

Now that we have constructed a decision tree where each tree leaf corresponds to a salient partition of the possible syntactic structures in the language, we then label these tree leaves as \textit{required-agreement} or \textit{chance-agreement}. For this we apply a threshold on the ratio of agreeing training samples within a leaf -- if the ratio exceeds a certain number the leaf will be judged as \emph{required-agreement}.
We experiment with two types of thresholds:
\paragraph{Hard Threshold:} We set a hard threshold on the ratio that is identical for all leaves. In all experiments, we set this threshold to~90\% based on manually inspecting some resulting trees to find a threshold that limited the number of non-agreeing syntactic structures being labeled as \emph{required-agreement}. 

\paragraph{Statistical Threshold:} \label{soft} 
Leaves with very few examples may exceed the hard threshold purely by chance. In order to better determine whether the agreements are indeed due to a true pattern of required agreement, we devise a thresholding strategy based on significance testing.
For all agreement-majority leaves, we apply a chi-squared goodness of fit test to compare the observed output distribution with an expected probability distribution specified by a null hypothesis.
Our null hypothesis $H_0$ will be that any agreement we observe is due to chance. If we reject the null hypothesis, we will conclude from the alternate hypothesis $H_1$ that there exists a grammatical rule requiring agreement for this leaf's cases:
\begin{align*}
  H_0&: \text{The leaf has \textit{chance-agreement}.}  \\
  H_1&: \text{The leaf has \textit{required-agreement}.}
\end{align*}
\noindent If there is no rule requiring agreement, we assume that the morphological properties of the head and the dependent token are independent and identically distributed discrete random variables following a categorical distribution. We compute the probability of chance agreement based on the number of values that the specific morphological property $f$ can take. 
Since morphological feature values are not equally probable, we use a probability proportional to the \textit{observed} value counts.
For a binary number property where 90\% of all observed occurrences are singular and 10\% are plural, the probability of chance agreement is equal to $0.82$=$0.9\!\times\!0.9$+$0.1\!\times\!0.1$, which gives the observed output distribution $p\!=\![0.18,0.82]$. Using $p$ we compute the expected frequency count $E_i = np_i$  where $n$ is the total number of samples in the given leaf,  $i\!=\![0,1]$ is  the output class of the leaf, 
and $p_i$ is the hypothesized proportion of observations for class $i$. The chi-squared test calculates the test statistic $\chi^2 $ as follows:
\[ \chi^2 = \sum_{i\in [0,1]} \frac{(O_i - E_i)^2}{E_i}\] where $O_i$ is the observed frequency count in the given leaf. The test outputs a $p$-value, which is the probability of observing a sample statistic as extreme as the test statistic. 
If the $p$-value is smaller than a chosen significance level (we use $0.01$) we reject the null hypothesis and label the leaf as \textit{required-agreement}.

The chi-squared test especially helps in being cautious with leaves with very few examples. However, for leaves with larger number of examples statistical significance alone is insufficient, because there are a large number of cases where there are small but significant differences from the ratio of agreement expected by chance.\footnote{One limitation of this is that rules that show agreement \emph{sometimes} get incorrectly labeled as either \emph{chance-agreement} or \emph{required-agreement}. We consider this in evaluation, but predicting sometimes agreement is relegated to future work.}
Therefore, in addition to comparing the $p$-value we also compute the \emph{effect size} which provides a quantitative measure on the magnitude of an effect \cite{sullivan2012using}. Cramér's phi $\phi_{c}$ \cite{cramir1946mathematical} is a commonly used method to measure the \emph{effect size}:
\[ \phi_{c} = \frac{\chi^2}{N(k-1)} \]\noindent where $\chi^2$ is the test statistic computed from the chi-squared test, $N$ is the total number of samples within a leaf, and $k$ is the degree of freedom (which in this case is $2$ since we have two output classes). \citet{cohen-1988-book} provides rules of thumb for interpreting these effect size. For instance, $\phi_c > 0.5$ is considered to be a large effect size and a large effect size suggests that the difference between the two hypotheses is important. Therefore, a leaf is labeled as \textit{required-agreement} when the $p$-value is less than the significance value and the effect size is greater than $0.5$. Now Leaf-1 in Figure \ref{fig:tree}(b) is correctly identified as \textit{chance-agreement}.

\paragraph{Rule Merging:}
Because we are aiming to have a concise, human-readable representation of agreement rules of a language, after labeling the tree leaves we merge sibling leaves with the same label as shown in Figure \ref{fig:tree}(c). Further, we collapse tree nodes having all leaves with the same label thereby reducing the apparent depth of the tree.

\section{Experimental Settings and Evaluation}
\label{experiment}
Our experiments aim to answer the following research questions: (1) can our framework extract linguistically plausible agreement rules across diverse languages? and (2) can it do so even if gold-standard syntactic analyses are not available? 
To answer the first question we evaluate rules extracted from gold-standard syntactic analysis (Sec.~\S\ref{oracle}). For the second question we experiment in low-resource and zero-shot scenarios using cross-lingual transfer to obtain parsers on the languages of interest, and evaluate the effect of noisy parsing results on the quality of rules (Sec.~\S\ref{sec:lowzero}).

\subsection{Settings}

\paragraph{\textbf{Data}}
We use the Surface-Syntactic Universal Dependencies (SUD) treebanks \cite{gerdes-etal-2018-sud,gerdes-etal-2019-improving} as the gold-standard source of complete syntactic analysis. The SUD treebanks are derived from Universal Dependencies (UD) \cite{nivre2016universal,nivre2018universal}, but unlike the UD treebanks which favor content words as heads, the SUD ones express dependency labels and links using purely syntactic criteria, which is more conducive to our goal of learning syntactic rules. We use the tool of \citet{gerdes-etal-2019-improving} to convert UD v.2.5 \citep{nivre-EtAl:2020:LREC} into SUD.  We only use the training portion of the treebanks for learning our rules.

\paragraph{\textbf{Rule Learning}}
We use \texttt{sklearn}'s \cite{scikit-learn} implementation of decision trees and train a separate model for each morphological feature $f$ for a given language. We experiment with six morphological features (Gender, Person, Number, Mood, Case, Tense) which are most frequently present across several languages. We perform a grid search over the decision tree parameters (detailed in Appendix A.1) and select the model performing best on the validation set. 
We report results with the \textit{Statistical Threshold} because on manual inspection we find the trees to be more reliable than the ones learnt from the \textit{Hard Threshold} (see Appendix A.5 for an example). 

\subsection{Evaluation}
\label{evaluation}
We explore two approaches to evaluate the extracted rules, one based on expert annotations, and an automated proxy evaluation.

\paragraph{\textbf{Expert Evaluation}}
\label{lingannotation}
Ideally, we would collect annotations for all head-relation-dependent triples in a treebank, but this would involve annotating hundreds of triples, requiring a large time commitment from linguists in each language we wish to evaluate.
Instead, for each language/treebank we extract and evaluate the top 20 most frequent ``head POS, dependency relation, dependent POS'' triples for the six morphological features amounting to 120 sets of triples to be annotated.%
\footnote{The top 20 most frequent triples covered approximately 95\% of the triples where this feature was active on average.}
We then present these triples with 10 randomly selected illustrative examples and ask a linguist to annotate whether there is a rule in this language governing agreement between the head-dependent pair for this relation. The allowed labels are: \textit{Almost always agree} if the construction must almost always exhibit agreement on the given feature; \textit{Sometimes agree} if the linked arguments sometimes must agree, but sometimes do not have to; \textit{Need not agree} if any agreement on the feature is random. An example of the annotation interface is shown in the Appendix A.2.

For each of the human annotated triples in feature $f$, we extract the label assigned to it by the learnt decision tree $\mathcal{T}$.  We find the leaf to which the given triple $t$ belongs and assign that leaf's label to the triple, referred  by $l_{\text{tree},f,t}$.  The human evaluation score (HS) for each triple marking feature $f$ is given by:\\
\[ 
\text{HS}_{f,t} = \mathbbm{1} \left\{
    \begin{array}{ll}
        1 & l_{\text{human},f,t} = l_{\text{tree},f,t} \\
        0 & \text{otherwise}
    \end{array}
\right.  \]
where $l_{\text{human},f,t}$ is the label assigned to the triple $t$ by the human annotator. These scores are then averaged across all annotated triples $T_f$ to get the human evaluation metric ($\text{HRM}$) for feature $f$ \[\text{HRM}_f = \frac{\sum_{t \in T_f} \text{HS}_{f,t}}{|T_f|}.\]

\paragraph{Automated Evaluation} 
As an alternative to the infeasible manual evaluation of all rules in every language, 
we propose an \textit{automated rule metric} (ARM) that evaluates how well the rules extracted from decision tree $\mathcal{T}$ fit to unseen gold-annotated test data.  For each triple $t$ marking feature $f$, we first retrieve all examples from the test data corresponding to that triple. Next, we calculate the empirical agreement by counting the fraction of test samples that exhibit agreement,  referred by $q_{f,t}$. 
For a \textit{required-agreement} leaf, we expect most test samples  satisfying that rule to show agreement.\footnote{There are exceptions: e.g.~when the head of dependent is a multiword expression (MWE), in which case dependency parsers might miss or pick only one of its constituents as head/dependent, or if the MWE is syntactically idiosyncratic.} To account for any exceptions to the rule and/or parsing-related errors, we use a threshold that acts as proxy for evaluating whether the given triple denotes \emph{required agreement}. We use a threshold of $0.95$, and if $q_{f,t} > 0.95$ then we assign the test label $l_{\text{test},f,t}$ for that triple as \textit{required-agreement}, and otherwise choose \textit{chance-agreement}.\footnote{We keep a 5\% margin to account for any exceptions or parsing errors based on the feedback given by the annotators.} Similar to the human evaluation, we compute a score for each triple $t$ marking feature $f$
\[ 
 \text{AS}_{t} = \mathbbm{1} \left\{
    \begin{array}{ll}
        1 & l_{\text{test},f,t} = l_{\text{tree},f,t} \\
        0 & \text{otherwise}
    \end{array}
\right. \]
 then average scores across all annotated triples in $T_f$ to get the ARM score for each feature $f$: \[\text{ARM}_f = \frac{\sum_{t \in T_f} \text{AS}_t}{|T_f|}\] 
\section{Experiments with Gold-Standard Data}
\label{oracle}
In this section, we evaluate the quality of the rules induced by our framework, using gold-standard syntactic analyses and learning the decision trees over triples obtained from the training portion of all SUD treebanks. As baseline, we compare with trees predicting all leaves as \textit{chance-agreement}.

 \begin{figure}
 \begin{tikzpicture}[trim left=-0.6cm,trim right=0cm]
    \begin{axis}[
            ybar,
            every axis plot post/.style={/pgf/number format/fixed},
            bar width=.25cm,
            width=8.5cm,
            height=3cm,
            ymajorgrids=false,
            yminorgrids=false,
            xtick={Gender, Case, Person},
            symbolic x coords={Gender, Case, Person},
            every x tick label/.append style={font=\small},
            every y tick label/.append style={font=\tiny},
            tick pos=left,
            axis x line*=bottom,
            axis y line*=left,
           title={\small \textcolor{blue}{$\blacksquare$} Indo-Aryan   \textcolor{ForestGreen}{$\blacksquare$} Uralic \textcolor{Dandelion}{$\blacksquare$} Slavic \\ \small \textcolor{RubineRed}{$\blacksquare$} Germanic \textcolor{Orchid}{$\blacksquare$} Baltic
           \textcolor{Brown}{$\blacksquare$} Semitic},
            title style={yshift=-0.2cm,align=left},
            ymin=-0.1,ymax=0.3,
            ytick={-0.1,0,0.1,0.2,0.3},
            ylabel shift={-.15cm},
            ylabel near ticks,
            ylabel={\small $\Delta$\textsc{ARM}},
            enlarge x limits=0.2,
        ]
        \addplot [style={blue,fill=blue,bar shift=-.775cm}] coordinates {(Gender,0.05344786533)};
        \addplot [style={blue,fill=blue,bar shift=-.775cm}] coordinates {(Case,0.0598063829)};
        \addplot [style={blue,fill=blue,bar shift=-.775cm}] coordinates {(Person,0.1501976285)};
       
        \addplot [style={ForestGreen,fill=ForestGreen,bar shift=-.475cm}] coordinates {(Case,	-0.03267807412)};
        \addplot [style={ForestGreen,fill=ForestGreen,bar shift=-.475cm}] coordinates {(Person,-0.0273548387)};
        
        \addplot [style={Dandelion,fill=Dandelion,bar shift=-.175cm}] coordinates {(Gender,0.1243908092)};
        \addplot [style={Dandelion,fill=Dandelion,bar shift=-.175cm}] coordinates {(Case,0.1512939509)};
        \addplot [style={Dandelion,fill=Dandelion,bar shift=-.175cm}] coordinates {(Person,	0.08224810833)};
       
        \addplot [style={RubineRed,fill=RubineRed,bar shift=0.175cm}] coordinates {(Gender,0.008743793217)};
        \addplot [style={RubineRed,fill=RubineRed,bar shift=0.175cm}] coordinates {(Case,0.2250138458)};
        \addplot [style={RubineRed,fill=RubineRed,bar shift=0.175cm}] coordinates {(Person,0.07782738095)};
        
        \addplot [style={Orchid,fill=Orchid,bar shift=.475cm}] coordinates {(Gender,0.14994517549)};
        \addplot [style={Orchid,fill=Orchid,bar shift=.475cm}] coordinates {(Case,0.161242511	)};
        \addplot [style={Orchid,fill=Orchid,bar shift=.475cm}] coordinates {(Person,0.08479532167)};
        
        \addplot [style={Brown,fill=Brown,bar shift=0.775cm}] coordinates {(Gender,0.0046382714)};
        \addplot [style={Brown,fill=Brown,bar shift=0.775cm}] coordinates {(Case,	0.1657217104)};
        \addplot [style={Brown,fill=Brown,bar shift=0.775cm}] coordinates {(Person,-0.01754385963)};
        
    \end{axis}
\end{tikzpicture}

\vspace{-1em}

\begin{tikzpicture}[trim left=-0.6cm,trim right=0cm]
    \begin{axis}[
            ybar,
            every axis plot post/.style={/pgf/number format/fixed},
            bar width=.25cm,
            width=8.5cm,
            height=3cm,
            ymajorgrids=false,
            yminorgrids=false,
            xtick={Number, Mood, Tense},
            symbolic x coords={Number, Mood, Tense},
            every x tick label/.append style={font=\small},
            every y tick label/.append style={font=\tiny},
            tick pos=left,
            axis x line*=bottom,
            axis y line*=left,
            ymin=-0.1,ymax=0.3,
            ytick={-0.1,0,0.1,0.2,0.3},
            ylabel shift={-.35cm},
            ylabel near ticks,
            ylabel={\small $\Delta$\textsc{ARM}},
            enlarge x limits=0.2,
        ]
        \addplot [style={blue,fill=blue,bar shift=-.775cm}] coordinates {(Number,0.0505050505)};
        \addplot [style={blue,fill=blue,bar shift=-.775cm}] coordinates {(Mood,0.6190476191)};
        \addplot [style={blue,fill=blue,bar shift=-.775cm}] coordinates {(Tense,-0.08333333335)};
        
        \addplot [style={Dandelion,fill=Dandelion,bar shift=-.175cm}] coordinates {(Number,0.12953421695)};
        \addplot [style={Dandelion,fill=Dandelion,bar shift=-.175cm}] coordinates {(Mood,-0.01614035088)};
        \addplot [style={Dandelion,fill=Dandelion,bar shift=-.175cm}] coordinates {(Tense,-0.05308025306)};

       \addplot [style={ForestGreen,fill=ForestGreen,bar shift=-.475cm}] coordinates {(Number, 0.04538198627)};
        \addplot [style={ForestGreen,fill=ForestGreen,bar shift=-.475cm}] coordinates {(Mood,-0.007936507933)};
        \addplot [style={ForestGreen,fill=ForestGreen,bar shift=-.475cm}] coordinates {(Tense,-0.012170285853)};

        \addplot [style={RubineRed,fill=RubineRed,bar shift=.175cm}] coordinates {(Number, 0.06242147099)};
        \addplot [style={RubineRed,fill=RubineRed,bar shift=.175cm}] coordinates {(Mood,0.06123737375)};
        \addplot [style={RubineRed,fill=RubineRed,bar shift=.175cm}] coordinates {(Tense,0.02933195592)};
        		
        \addplot [style={Orchid,fill=Orchid,bar shift=.475cm}] coordinates {(Number,0.17255645476)};
        \addplot [style={Orchid,fill=Orchid,bar shift=.475cm}] coordinates {(Mood,0)};
        \addplot [style={Orchid,fill=Orchid,bar shift=.475cm}] coordinates {(Tense,-0.1216931217)};

        \addplot [style={Brown,fill=Brown,bar shift=0.775cm}] coordinates {(Number,	0.05691056913)};
        \addplot [style={Brown,fill=Brown,bar shift=0.775cm}] coordinates {(Mood,	0	)};
        \addplot [style={Brown,fill=Brown,bar shift=0.775cm}] coordinates {(Tense,0.2	)};
        	
    \end{axis}
\end{tikzpicture}
\vspace{-.5em}
\vspace{-.5em}
 \caption{Difference in the ARM scores of decision trees over gold-standard syntactic analysis with baseline trees where all leaves predict \textit{chance-agreement}.}
 \label{fig:langfamily}
 \end{figure}
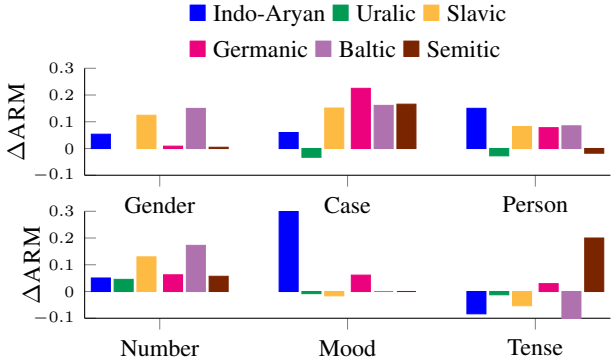
 
 \begin{figure}
\begin{tikzpicture}[trim left=-0.6cm,trim right=0cm]
    \begin{axis}[
            ybar,
            every axis plot post/.style={/pgf/number format/fixed},
            bar width=.05cm,
            width=9cm,
            height=3.7cm,
            ymajorgrids=false,
            yminorgrids=false,
            xtick={Gender, Case, Person, Tense},
            symbolic x coords={Gender, Case, Person, Tense},
            every x tick label/.append style={font=\small},
            every y tick label/.append style={font=\tiny},
            tick pos=left,
            axis x line*=bottom,
            axis y line*=left,
            title={\small \textcolor{blue}{$\blacksquare$} Hindi \textcolor{Dandelion}{$\blacksquare$} Russian \textcolor{ForestGreen}{$\blacksquare$} North Sami \textcolor{RubineRed}{$\blacksquare$} Tamil \textcolor{Orchid}{$\blacksquare$} Arabic},
            title style={yshift=-.5cm},
            ymin=0,ymax=1.1,
            ytick={0,.2,.4,.6,.8,1},
            ylabel shift={-.15cm},
            ylabel near ticks,
            ylabel={\small \textsc{arm}},
            enlarge x limits=0.2,
        ]
        \addplot [style={blue,fill=blue,opacity=0.4,bar shift=-.4cm,bar width=.15cm}] coordinates {(Gender,0.5859375)};
        \addplot [style={blue,fill=blue,opacity=0.4,bar shift=-.4cm,bar width=.15cm}] coordinates {(Case,0.6535433071)};
        \addplot [style={blue,fill=blue,opacity=0.4,bar shift=-.4cm, bar width=.15cm}] coordinates {(Person,0.04545454545)};
        \addplot [style={blue,fill=blue,opacity=0.4,bar shift=-.4cm,bar width=.15cm}] coordinates {(Tense,0.3333333333)};
        
        \addplot [style={blue,fill=blue,opacity=1,bar shift=-.4cm}] coordinates {(Gender,0.5)};
        \addplot [style={blue,fill=blue,opacity=1,,bar shift=-.4cm}] coordinates {(Case,0.6299212598)};
        \addplot [style={blue,fill=blue,opacity=1,,bar shift=-.4cm}] coordinates {(Person,0.04545454545)};
        \addplot [style={blue,fill=blue,opacity=1,,bar shift=-.4cm}] coordinates {(Tense,0.3333333333)};
        
        \addplot [style={Dandelion,fill=Dandelion,opacity=0.4,bar shift=-.2cm,bar width=.15cm}] coordinates {(Gender,0.6968325792)};
        \addplot [style={Dandelion,fill=Dandelion,opacity=0.4,bar shift=-.2cm,bar width=.15cm}] coordinates {(Case,0.6494252874)};
        \addplot [style={Dandelion,fill=Dandelion,opacity=0.4,bar shift=-.2cm,bar width=.15cm}] coordinates {(Person,0.625)};
        \addplot [style={Dandelion,fill=Dandelion,opacity=0.4,bar shift=-.2cm,bar width=.15cm}] coordinates {(Tense,0.7272727273)};
        
        \addplot [style={Dandelion,fill=Dandelion,opacity=1,,bar shift=-.2cm}] coordinates {(Gender,0.6244343891)};
        \addplot [style={Dandelion,fill=Dandelion,opacity=1,,bar shift=-.2cm}] coordinates {(Case,	0.5747126437)};
        \addplot [style={Dandelion,fill=Dandelion,opacity=1,,bar shift=-.2cm}] coordinates {(Person,	0.6666666667)};
        \addplot [style={Dandelion,fill=Dandelion,opacity=1,,bar shift=-.2cm}] coordinates {(Tense,0.8181818182)};
        
        \addplot [style={ForestGreen,fill=ForestGreen,opacity=1,,bar shift=0cm}] coordinates {(Case,0.7407407407)};
        \addplot [style={ForestGreen,fill=ForestGreen,opacity=0.4,bar shift=0cm,bar width=.15cm}] coordinates {(Case,0.6666666667)};

        \addplot [style={ForestGreen,fill=ForestGreen,opacity=0.4,bar shift=0cm,bar width=.15cm}] coordinates {(Tense,0.45454545455)};
        \addplot [style={ForestGreen,fill=ForestGreen,opacity=1,,bar shift=0cm}] coordinates {(Tense,0.4545454545)};
        
        \addplot [style={RubineRed,fill=RubineRed,opacity=0.4,bar shift=.2cm,bar width=.15cm}] coordinates {(Gender,0.6818181818)};
        \addplot [style={RubineRed,fill=RubineRed,opacity=0.4,bar shift=.2cm,bar width=.15cm}] coordinates {(Case,0.8461538462)};
        \addplot [style={RubineRed,fill=RubineRed,opacity=0.4,bar shift=.2cm,bar width=.15cm}] coordinates {(Person,0.09090909091)};
        \addplot [style={RubineRed,fill=RubineRed,opacity=0.4,bar shift=.2cm,bar width=.15cm}] coordinates {(Tense,0.625)};
        
        \addplot [style={RubineRed,fill=RubineRed,opacity=0.4,opacity=1,,bar shift=.2cm}] coordinates {(Gender,	0.6590909091)};
        \addplot [style={RubineRed,fill=RubineRed,opacity=1,,bar shift=.2cm}] coordinates {(Case,0.6923076923)};
        \addplot [style={RubineRed,fill=RubineRed,opacity=1,,bar shift=.2cm}] coordinates {(Person,0.09090909091)};
        \addplot [style={RubineRed,fill=RubineRed,opacity=1,,bar shift=.2cm}] coordinates {(Tense,0.625)};
        

        \addplot [style={Orchid,fill=Orchid,opacity=0.4,bar shift=.4cm,bar width=.15cm}] coordinates {(Gender,0.6056338028)};
        \addplot [style={Orchid,fill=Orchid,opacity=0.4,bar shift=.4cm,bar width=.15cm}] coordinates {(Case,0.6265060241)};
        \addplot [style={Orchid,fill=Orchid,opacity=0.4,bar shift=.4cm,bar width=.15cm}] coordinates {(Person,0.46875)};

        \addplot [style={Orchid,fill=Orchid,opacity=1,,bar shift=.4cm}] coordinates {(Gender,0.7183098592)};
        \addplot [style={Orchid,fill=Orchid,opacity=1,,bar shift=.4cm}] coordinates {(Case,0.7469879518)};
        \addplot [style={Orchid,fill=Orchid,opacity=1,,bar shift=.4cm}] coordinates {(Person,0.46875)};

        \node[style={xshift=0cm}] at (axis cs:Gender,.05) {\textcolor{ForestGreen}{$\mathbf{\times}$}};
        \node[style={xshift=0cm}] at (axis cs:Person,.05) {\textcolor{ForestGreen}{$\mathbf{\times}$}};
        \node[style={xshift=.4cm}] at (axis cs:Tense,.05) {\textcolor{Orchid}{$\mathbf{\times}$}};
    \end{axis}
\end{tikzpicture}
\vspace{-1.7em}
\caption{ Our approach (shaded bars) outperforms the chance-agreement baseline (solid bars) in all cases where there exist agreement rules. Features not present in the language are marked with $\mathbf{\times}$.} 
\label{fig:oracle}
\end{figure}
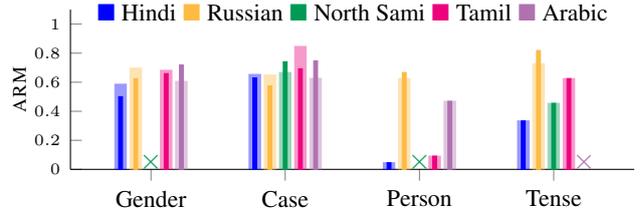

The extracted rules have an $0.574$ ARM score (averaged across all treebanks and features), outperforming the baseline scores by $0.074$ ARM points.\footnote{Individual scores for each treebank are in Appendix A.5.} Of all the 451 decision trees across all treebanks and features, we find 78\% trees outperforming the baseline trees. In Figure \ref{fig:langfamily}, we show the improvements over the baseline averaged across language families/genera. In families with extensive agreement systems such as Slavic and Baltic our models clearly outperform the baseline discovering correct rules, as they do for the other Indo-European genera, Indo-Aryan and Germanic.
For mood and tense, the \emph{chance-agreement} baseline performs on par with our method. This is not surprising because there is little agreement observed for these features given that only verbs and auxiliary verbs mark these features. We find that for both tense and mood in the Indo-Aryan family,  our model identifies  
\emph{required-agreement} primarily for conjoined verbs, which mostly need to agree only if they share the same subject. However, subsequent analysis revealed that in the treebanks nearly 50\% of the agreeing verbs do not share the same subject but do agree by chance. 


Agreement for Indo-European languages like Hindi and Russian is well documented~\cite{comrie1984reflections,crockett1976agreement} and is reflected in our large improvements over the baseline (Figure \ref{fig:oracle}). Similarly, Arabic exhibits extensive agreement on noun phrases including determiners and adjectives~\cite{aoun1994agreement}. We find that for Arabic gender the lower ARM scores of our method are an artifact of the small test data.

North Sami is an interesting test bed: as a Uralic language, case agreement would be somewhat unexpected and indeed our model's predictions are not better than the baseline. Nevertheless, with our interface we find patterns of rare positive paratactic constructions with required agreement where demonstrative pronouns overwhelmingly agree with their heads.\footnote{Leaf 3 here: \url{https://bit.ly/34mHTeG}}
The case decision tree also uncovers interesting patterns of~100\% agreement on Tamil constructions with nominalized verbs (Gerunds) where the markings propagate to the whole phrase.

\paragraph{Conciseness of Extracted Rules}
 \input{images/correlation_leaves}
We further analyze the decision trees learnt by our framework for conciseness and find that the trees grow more complex with increasing morphological complexity of languages as seen in Figure \ref{fig:morphologyrichness}. To compute the morphological complexity of a language, we use the word entropy measure proposed by \citet{bentz-etal-2016-comparison} which measures the average information content of words and is computed as follows:
\[H(D) = -\sum_{i \in V} p(w_i)\log p(w_i) \] where $V$ is the vocabulary, $D$ is the monolingual text extracted from the training portion of the respective treebank, $p(w_i)$ is the word type frequency normalized by the total tokens. Since this entropy doesn't account for unseen word types, \citet{bentz-etal-2016-comparison} use the \emph{James-Stein shrinkage} estimator \cite{hausser2009entropy} to calculate $p(w_i)$:
\[p(w_i) = \lambda p^{\text{target}}(w_i) + (1-\lambda)p^{\text{ML}}(w_i)\]
where $\lambda\!\in\![0,1]$, $p^{\text{target}}$ denotes the maximum entropy case given by the uniform distribution $\frac{1}{V}$ and $p^{\text{ML}}$ is the maximum likelihood estimator which is given by the normalized word type frequency. 
Languages with a larger word entropy are considered to be morphologically rich as they pack more information into the words. In Figure \ref{fig:morphologyrichness} we plot the morphological richness with the average number of leaves across all features and find these to be highly correlated.

\begin{figure}
    \pgfplotstableread[row sep=\\,col sep=&]{
Language & Gender & Number & Person & Tense & Case & Mood \\
Greek & 0.85 & 0.8 & 0.95 & 0.7647058824 & 1 & 1 \\
Russian & 0.8 & 0.9 & 0.7 & 0.8 & 0.7368421053 & 0 \\
Catalan & 0.8 & 0.9473684211 & 0.6 & 0.2222222222 & 0 & 0.7368421053 \\
}\annotationdata

\begin{tikzpicture}[trim left=-0.6cm,trim right=0cm]
    \begin{axis}[
            ybar,
            every axis plot post/.style={/pgf/number format/fixed},
            bar width=.24cm,
            width=8.5cm,
            height=3.5cm,
            ymajorgrids=false,
            yminorgrids=false,
            xtick={Greek,Russian,Catalan},
            symbolic x coords={Greek,Russian,Catalan},
            every x tick label/.append style={font=\small},
            every y tick label/.append style={font=\tiny},
            tick pos=left,
            axis x line*=bottom,
            axis y line*=left,
            title={\small \textcolor{blue}{$\blacksquare$} Gender \textcolor{Dandelion}{$\blacksquare$} Number \textcolor{ForestGreen}{$\blacksquare$} Person
            \small \textcolor{RubineRed}{$\blacksquare$} Tense \textcolor{Orchid}{$\blacksquare$} Case \textcolor{Brown}{$\blacksquare$} Mood},
            title style={yshift=-0.2cm,align=left},
            ymin=0,ymax=1.1,
            ytick={0,.2,.4,.6,.8,1},
            ylabel shift={-.15cm},
            ylabel near ticks,
            ylabel={\small \textsc{arm}},
            enlarge x limits=0.2,
        ]
        \addplot [style={blue,fill=blue}] table[x=Language,y=Gender]{\annotationdata};
        \addplot [style={Dandelion,fill=Dandelion}] table[x=Language,y=Number]{\annotationdata};
        \addplot [style={ForestGreen,fill=ForestGreen}] table[x=Language,y=Person]{\annotationdata};
        \addplot [style={RubineRed,fill=RubineRed}] table[x=Language,y=Tense]{\annotationdata};
        \addplot [style={Orchid,fill=Orchid}] table[x=Language,y=Case]{\annotationdata};
        \addplot [style={Brown,fill=Brown}] table[x=Language,y=Mood]{\annotationdata};
        \node[style={xshift=.7cm}] at (axis cs:Russian,.05) {\textcolor{Brown}{$\mathbf{\times}$}};
        \node[style={xshift=.45cm}] at (axis cs:Catalan,.05) {\textcolor{Orchid}{$\mathbf{\times}$}};
    \end{axis}
\end{tikzpicture}
\vspace{-.7cm}
 
    \caption{Annotation accuracy for Greek, Russian and Catalan per each morphological feature. }
    \label{fig:annotationresult}
    \vspace{-1em}
\end{figure}
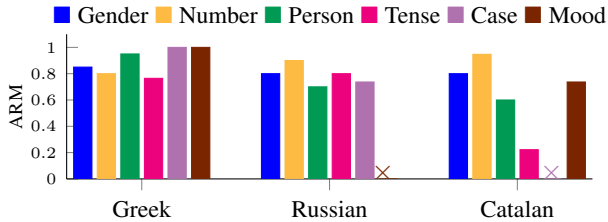

\paragraph{Manual Evaluation Results}
\label{lingannotation_}
We conduct an expert evaluation for Greek (el), Russian (ru) and Catalan (ca) as described in Section \S \ref{lingannotation}. For a strict setting, we consider both \textit{Sometimes agree} and \textit{Need not agree} as \textit{chance-agreement} and report the human evaluation metric (HRM) in Figure~\ref{fig:annotationresult}. Overall, our method extracts first-pass grammar rules achieving 89\% accuracy for Greek, 78\% for Russian and 66\% for Catalan. 

In most error cases, like person in Russian, our model produces \textit{required-agreement} labels, which we can attribute to skewed data statistics in the treebanks. In Russian and Greek, for instance, conjoined verbs only need to agree in person and number if they share the same subject (in which case they \textit{implicitly} agree because they both must agree with the same subject phrase). In the treebanks, though, only 15\% of the agreeing verbs do indeed share the same subject -- the rest agree by chance. In a reverse example from Catalan, the overwhelming majority (92\%) of 8650 tokens are in the third-person, causing our model to label all leaves as chance agreement despite the fact that person/number agreement is required in such cases. Similarly for tense in Catalan, our framework predicts \textit{chance-agreement} for auxiliary verbs with verbs as their dependent because of overwhelming majority of disagreeing examples. We believe this is because of both annotation artifact and the way past tense is realized.

To demonstrate how well the automated evaluation correlates with the human evaluation protocol, we compute the Pearson's correlation ($r$) between the ARM and HRM for each language under four model settings: \emph{simulate-50, simulate-100, baseline} and \emph{gold}. \emph{simulate-x} is a  simulated low-resource setting where the model is trained using $x$ gold-standard syntactically analysed data.\footnote{More details on the experimental setup in \S ~\ref{sec:lowzeroexpt}.} The \emph{baseline} setting is the one where all leaves predict \emph{chance-agrement} and under the \emph{gold} setting we train using the entire gold-standard data. We compute the ARM and HRM scores for the rules learnt under each of the four settings and report the Pearson's correlation, averaged across all features. Overall, we observe a moderate correlation for all three languages, with  $r=0.59$ for Greek, $r\!=\!0.41$ for Russian and $r\!=\!0.38$ for Catalan. The correlations are very strong for some features such as Gender ($r_{\text{el}}\!=\!0.97$, $r_{\text{ru}}\!=\!0.82$, $r_{\text{ca}}\!=\!0.98$) and Number ($r_{\text{el}}\!=\!0.97$, $r_{\text{ru}}\!=\!0.69$, $r_{\text{ca}}\!=\!0.96$) where we expect to see extensive agreement.


\section{Low-Resource Experiments}
\label{sec:lowzero}

\subsection{Simulated Zero-/Few-Shot Experiments}
\label{sec:lowzeroexpt}

It is not always possible to have access to gold-standard syntactic analyses. Therefore, in order to investigate how the quality of rules are affected by the quality of syntactic analysis, we conduct simulation experiments by varying the amount of gold-standard syntactically analysed training data. For each language, we sample $x$ fully parsed sentences from the its treebank out of $L$ training sentences available. For the remaining $L-x$ sentences, we use \emph{silver} syntactic analysis i.e., we train a syntactic analysis model on $x$ sentences and use the model predictions for the $L-x$ sentences.

 \paragraph{Data and Setup:}
We experiment with Spanish, Greek, Belarusian and Lithuanian. For transfer learning, we use Portuguese, Ancient Greek, Ukrainian and Latvian treebanks respectively. The data statistics and details are  in Appendix A.2. 

We train \texttt{Udify} \cite{kondratyuk-straka-2019-75}, a parser that jointly predict POS tags, morphological features, and dependency trees, using the $x$ gold-standard sentences as our training data. We generate model predictions on the remaining $L-x$ sentences. Finally, we concatenate the $x$ gold data with the $L-x$ automatically parsed data from which we extract the training data for learning the decision tree. We experiment with $x= [50, 100, 500]$ gold-standard sentences. To account of sampling randomness, we repeat the process 5 times and report averages across runs. 

To further improve the quality of the automatically obtained syntactic analysis, we use cross-lingual transfer learning where we train the \texttt{Udify} model by concatenating $x$ sentences of the target language with the entire treebank of the related language. 
 We also conduct zero-shot experiments under this setting where we directly use the \texttt{Udify} model trained only on the related language and get the model predictions on $L$ sentences. As before, we train five decision trees for each $x$ setting and report the average ARM over the test data. 

\begin{figure}
\pgfplotstableread[row sep=\\,col sep=&]{
Number & Greek & GreekAnc\\
25	& 0	& 0.625 \\
50	& 0.539 & 0.678\\
100	& 0.637 & 0.641 \\
500	& 0.68 & 0.686 \\
1662 & 0.6272727273 & 0.6272727273\\
}\greekdata
\pgfplotstableread[row sep=\\,col sep=&]{
Number & Spa & SpaPor \\
25	& 0	& 0.632 \\
50	& 0.611 & 0.627 \\
100 &	0.602 &	0.652 \\
500 &	0.611 &	0.659 \\
14187	 & 0.6439393939	& 0.6439393939 \\
}\spadata
\pgfplotstableread[row sep=\\,col sep=&]{
Number & Bel & BelUkr \\
25 & 0 &	0.592 \\
50 & 0.593 &	0.63 \\
100 & 0.564 &	0.606 \\
319 & 0.6461538462 & 0.6461538462 \\
}\beldata
\pgfplotstableread[row sep=\\,col sep=&]{
Number & Lit & LitLav \\
25 &	0 &	0.663 \\
50 &	0.611 & 0.618 \\
100	& 0.602 &	0.632 \\
500 &	0.611& 	0.62 \\
2341 &	0.625 &	0.625 \\
}\litdata

\begin{tikzpicture}[trim left=-0.85cm,trim right=0cm]
    \begin{axis}[
            every axis plot post/.style={/pgf/number format/fixed},
            width=4.5cm,
            height=3cm,
            ymajorgrids=false,
            yminorgrids=false,
            legend style={draw=none},
            xtick={25,50,100,500,1662},
            xticklabels={0,50,100,500,1662},
            every x tick label/.append style={font=\tiny},
            every y tick label/.append style={font=\tiny},
            tick pos=left,
            axis y line*=left,
            axis x line*=bottom,
            xmode=log,
            log ticks with fixed point,
            title={{Greek\\(Ancient Greek)}},
            title style={yshift=-.2cm,font=\small,align=center},
            ymin=0.4,ymax=1,
            ylabel shift={-.2cm},
            ylabel near ticks,
            ylabel={ARM},
            ylabel style={font=\small},
        ]
        \addplot [style={blue},mark=o,] table[x=Number,y=Greek]{\greekdata};
        \addplot [style={red},mark=x,nodes near coords,nodes near coords align={south},every node near coord/.append style={font=\tiny,color=red},] table[x=Number,y=GreekAnc]{\greekdata};
        \node [blue, left] at (axis cs: 50,.53)  {\tiny 0.53};
        \node [blue, below] at (axis cs: 100,.63) {\tiny 0.63};
        \node [blue, below] at (axis cs: 500,.68)  {\tiny 0.68};
        \node [blue, below] at (axis cs: 1662,.62) {\tiny 0.62};
    \end{axis}
\end{tikzpicture}
\begin{tikzpicture}[trim left=-3.7cm,trim right=0cm]
    \begin{axis}[
            every axis plot post/.style={/pgf/number format/fixed},
            width=4.5cm,
            height=3cm,
            ymajorgrids=false,
            yminorgrids=false,
            legend style={draw=none},
            xtick={25,50,100,500,2341},
            xticklabels={0,50,100,500,2341},
            every x tick label/.append style={font=\tiny},
            every y tick label/.append style={font=\tiny},
            tick pos=left,
            axis y line*=left,
            axis x line*=bottom,
            xmode=log,
            log ticks with fixed point,
            title={{Lithuanian\\(Latvian)}},
            title style={yshift=-.2cm,font=\small,align=center},
            ymin=0.5,ymax=0.7,
            ylabel shift={-1cm},
             tick={0.5,0.6,0.7},
        ]
        \addplot [style={blue},mark=o,nodes near coords,nodes near coords align={north},every node near coord/.append style={font=\tiny,color=blue},] table[x=Number,y=Lit]{\litdata};
        \addplot [style={red},mark=x,] table[x=Number,y=LitLav]{\litdata};
        \node [red, above] at (axis cs: 25,.663)  {\tiny 0.663};
        \node [red, above] at (axis cs: 50,.618)  {\tiny 0.618};
        \node [red, above] at (axis cs: 100,.632) {\tiny 0.632};
        \node [red, above] at (axis cs: 500,.62)  {\tiny 0.62};
        \node [red, above] at (axis cs: 2341,.625) {\tiny 0.625};
    \end{axis}
\end{tikzpicture}

\begin{tikzpicture}[trim left=-0.85cm,trim right=0cm]
    \begin{axis}[
            every axis plot post/.style={/pgf/number format/fixed},
            width=5.5cm,
            height=3cm,
            ymajorgrids=false,
            yminorgrids=false,
            legend style={draw=none,at={(1.2,1.7)},font=\tiny},
            xtick={25,50,100,500,14187},
            xticklabels={0,50,100,500,14187},
            every x tick label/.append style={font=\tiny},
            every y tick label/.append style={font=\tiny},
            tick pos=left,
            axis y line*=left,
            axis x line*=bottom,
            xmode=log,
            log ticks with fixed point,
            title={{Spanish\\(Portuguese)}},
            title style={yshift=-.2cm,xshift=-0.5cm,font=\small,align=center},
            ymin=0.5,ymax=0.7,
            ylabel shift={-.2cm},
            ylabel near ticks,
            ylabel={ARM},
            ylabel style={font=\small},
            ytick={0.5,0.6,0.7},
        ]
        \addplot [style={blue},mark=o,nodes near coords,nodes near coords align={north east},every node near coord/.append style={font=\tiny,color=blue},] table[x=Number,y=Spa]{\spadata};
        \addplot [style={red},mark=x,] table[x=Number,y=SpaPor]{\spadata};
        \node [red, above] at (axis cs: 20,.63)  {\tiny 0.63};
        \node [red, above] at (axis cs: 45,.62)  {\tiny 0.62};
        \node [red, above] at (axis cs: 105,.65) {\tiny 0.65};
        \node [red, above] at (axis cs: 500,.65)  {\tiny 0.65};
        \node [red, above] at (axis cs: 14187,.64) {\tiny 0.64};
        \legend{without,with transfer}
    \end{axis}
\end{tikzpicture}
\begin{tikzpicture}[trim left=-4.5cm,trim right=0cm]
    \begin{axis}[
            every axis plot post/.style={/pgf/number format/fixed},
            width=3.5cm,
            height=3cm,
            ymajorgrids=false,
            yminorgrids=false,
            legend style={draw=none},
            xtick={25,50,100,319},
            xticklabels={0,50,100,319},
            every x tick label/.append style={font=\tiny},
            every y tick label/.append style={font=\tiny},
            tick pos=left,
            axis y line*=left,
            axis x line*=bottom,
            xmode=log,
            log ticks with fixed point,
            title={{Belarusian\\(Ukrainian)}},
            title style={yshift=-.2cm,font=\small,align=center},
            ymin=0.5,ymax=0.7,
            ylabel shift={-1cm},
            ytick={0.5,0.6,0.7},
        ]
        \addplot [style={blue},mark=o,] table[x=Number,y=Bel]{\beldata};
        \addplot [style={red},mark=x,] table[x=Number,y=BelUkr]{\beldata};
        \node [red, above] at (axis cs: 25,.59)  {\tiny 0.59};
        \node [red, above] at (axis cs: 50,.63)  {\tiny 0.63};
        \node [red, above] at (axis cs: 100,.60) {\tiny 0.60};
        \node [red, above] at (axis cs: 311,.64)  {\tiny 0.64};
        \node [blue, left] at (axis cs: 50,.59) {\tiny 0.59};
        \node [blue, below] at (axis cs: 100,.56) {\tiny 0.56};
        \node [blue, below] at (axis cs: 311,.64) {\tiny 0.64};
    \end{axis}
\end{tikzpicture}
\vspace{-3mm}
    \caption{ Comparing the (avg.) ARM score for \texttt{Number} agreement with and without cross-lingual transfer learning (transfer language in parenthesis). $x$-axis in log space. The higher the ARM the better. }
    \vspace{-3mm}
    \label{fig:my_label}
\end{figure}
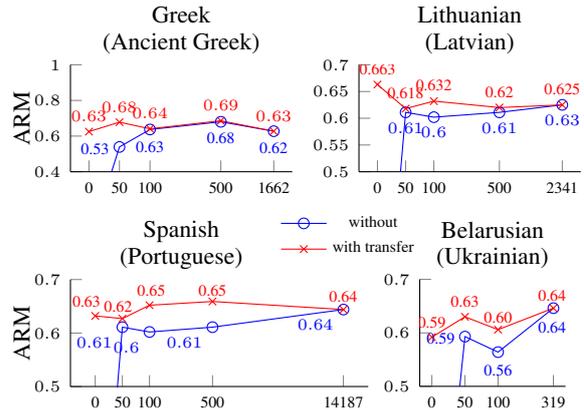

\paragraph{Results}
We report the results for Number agreement in Figure \ref{fig:my_label}. Similar plots for other languages and features can be found in the Appendix A.5. 
We observe that using cross-lingual transfer learning (CLTL) already leads to high scores across all languages even in zero-shot settings where we do not use any data from the gold-standard treebank. 
Taking Spanish gender as an example, 93\% of the rule-triples extracted from the gold-standard tree (which are overwhelmingly correct) are also extracted by the zero-shot tree. The zero-shot tree only makes a few mistakes (shown in Table~\ref{tab:es_zeroshot} and reflected in its overall ARM score) on certain proper noun and auxiliary verb constructions.
\begin{table}[t]
    \begin{center}
    \resizebox{\columnwidth}{!}{
    \begin{tabular}{@{}c|c|cc@{}}
    [Relation, Head, Dependent] & correct label & gold & zero-shot \\
    \midrule
        det, \textsc{noun}, \textsc{det.} & almost always & required & required  \\
        mod, \textsc{noun}, \textsc{adj} & almost always & required & required \\
        flat, \textsc{propn}, \textsc{propn} & almost always & required & \textcolor{red}{chance} \\
        mod, \textsc{propn}, \textsc{propn} & almost always & required & \textcolor{red}{chance} \\
        appos, \textsc{propn}, \textsc{propn} & sometimes & required & chance \\
        comp:aux@pass, \textsc{aux}, \textsc{verb} &  need not  & chance & \textcolor{red}{required} \\
        conj, \textsc{propn}, \textsc{propn} & need not & required & \textcolor{red}{chance} \\
    \multicolumn{2}{r}{ARM score over the test set:} & 0.644 & 0.632\\
    \bottomrule
    \end{tabular}
    }
    \caption{The Spanish gender rules extracted in a zero-shot setting are generally similar to the ones extracted from the gold data (93\%). We \textcolor{red}{highlight} the few mistakes that the zero-shot tree makes.}
   \label{tab:es_zeroshot}
     \end{center}
\end{table}
Interestingly, using CLTL, training with just 50 gold-standard target language sentences is almost equivalent to training with 100 or 500 gold-standard sentences. This opens new avenues for language documentation: with as few as 50 expertly-annotated syntactic analysis of a new language and CLTL our framework can produce decent first-pass agreement rules. Needless to say, in most cases the extracted rules improve as we increase the number of gold-standard sentences and CLTL further helps bridge the data availability gap for low-resource settings. 


\subsection{Real Zero-Shot Experiments}
Some languages like Breton, Buryat, Faroese, Tagalog and Welsh have test data only; there is no gold-standard training data available, which presents a \emph{true} zero-shot setting. In such cases, we can still extract grammar rules with our framework using zero-shot dependency parsing.

\paragraph{Data and Setup:}
We collect raw text for the above languages from the Leipzig corpora~\cite{goldhahn2012building}. Data statistics are listed in Appendix A.2. We parse these sentences using the ``universal" \texttt{Udify} model that has been pre-trained on all of the UD treebanks, as released by \cite{kondratyuk-straka-2019-75}. As before, we use these automatically parsed syntactic analyses to extract the rules which we evaluate with ARM over the gold standard test data of the corresponding SUD treebanks.

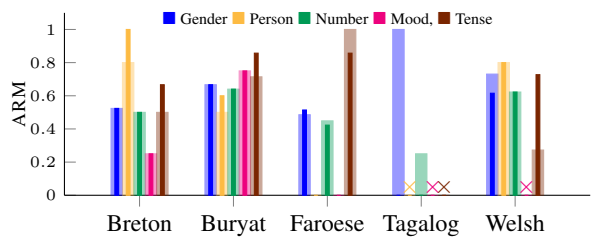
\begin{figure}[t]
    \small
    \pgfplotstableread[row sep=\\,col sep=&]{
Language &	Gender &	Person &	Number &	Mood &	Tense & BaseGender  & BasePerson &	BaseNumber&		BaseMood & BaseTense\\
Breton & 0.5238095238 & 0.8 & 0.5 & 0.25 & 0.5 & 0.5238095238 & 1 & 0.5 & 0.25 & 0.6666666667 \\
 Buryat & 0.6666666667 & 0.5 & 0.64 & 0.75 & 0.7142857143 & 0.6666666667 & 0.6 & 0.64 & 0.75 & 0.8571428571 \\
 Faroese	&0.4857142857 & 0 & 0.4487179487 & 0 & 1 & 0.5142857143 & 0 & 0.4230769231 & 0 & 0.8571428571\\
 Tagalog	& 1	& 0	& 0.25	& 0	& 0  &0 &	0&	0&	0&	0 \\
 Welsh	&0.7307692308 & 0.8 & 0.6226415094 & 0.0 & 0.2727272727 & 0.6153846154 & 0.8 & 0.6226415094 & 0.0 & 0.7272727273\\
}\zeroshotdata

\begin{tikzpicture}[trim left=-0.6cm,trim right=0cm]
    \begin{axis}[
            ybar,
            every axis plot post/.style={/pgf/number format/fixed},
           bar width=.05cm,
            width=8.5cm,
            height=4cm,
            ymajorgrids=false,
            yminorgrids=false,
            xtick={Breton,Buryat,Faroese, Tagalog, Welsh},
            symbolic x coords={Breton,Buryat,Faroese, Tagalog, Welsh},
            every x tick label/.append style={font=\small},
            every y tick label/.append style={font=\tiny},
            tick pos=left,
            axis x line*=bottom,
            axis y line*=left,
            title={\tiny \textcolor{blue}{$\blacksquare$} Gender  \textcolor{Dandelion}{$\blacksquare$} Person
            \textcolor{ForestGreen}{$\blacksquare$} Number
            \textcolor{RubineRed}{$\blacksquare$} Mood,
            \textcolor{Brown}{$\blacksquare$} Tense},  
            title style={yshift=-.5cm,align=left},
            ymin=0,ymax=1.1,
            ytick={0,.2,.4,.6,.8,1},
            ylabel shift={-.15cm},
            ylabel near ticks,
            ylabel={\small \textsc{arm}},
            enlarge x limits=0.2,
        ]
        \addplot [style={blue,fill=blue,opacity=0.4,bar shift=-0.3cm, bar width=.15cm}] table[x=Language,y=Gender]{\zeroshotdata};
        \addplot [style={blue,fill=blue,bar shift=-0.3cm}] table[x=Language,y=BaseGender]{\zeroshotdata};
        \addplot [style={Dandelion,fill=Dandelion,opacity=0.4,bar shift=-0.15cm,bar width=.15cm}] table[x=Language,y=Person]{\zeroshotdata};
        \addplot [style={Dandelion,fill=Dandelion,bar shift=-0.15cm}] table[x=Language,y=BasePerson]{\zeroshotdata};
         \addplot [style={ForestGreen,fill=ForestGreen,opacity=0.4,bar shift=0cm,bar width=.15cm}] table[x=Language,y=Number]{\zeroshotdata};
         \addplot [style={ForestGreen,fill=ForestGreen,bar shift=0cm}] table[x=Language,y=BaseNumber]{\zeroshotdata};
        \addplot [style={RubineRed,fill=RubineRed,opacity=0.4,bar shift=0.15cm,bar width=.15cm}] table[x=Language,y=Mood]{\zeroshotdata};
         \addplot [style={RubineRed,fill=RubineRed,bar shift=0.15cm}] table[x=Language,y=BaseMood]{\zeroshotdata};
         \addplot [style={Brown,fill=Brown,opacity=0.4,bar shift=0.3cm,bar width=.15cm}] table[x=Language,y=Tense]{\zeroshotdata};
         \addplot [style={Brown,fill=Brown,bar shift=0.3cm}] table[x=Language,y=BaseTense]{\zeroshotdata};
         %
        \node[style={xshift=-.15cm}] at (axis cs:Tagalog,.05) {\textcolor{Dandelion}{$\times$}};
         \node[style={xshift=.15cm}] at (axis cs:Tagalog,.05) {\textcolor{RubineRed}{$\times$}};
         \node[style={xshift=.3cm}] at (axis cs:Tagalog,.05) {\textcolor{Brown}{$\times$}};
         \node[style={xshift=.15cm}] at (axis cs:Welsh,.05) {\textcolor{RubineRed}{$\mathbf{\times}$}};
    \end{axis}
\end{tikzpicture}
\vspace{-.3cm}
 
    \caption{In most cases our framework (shaded bars) extracts a good first-pass specification for \textit{true} zero-shot settings. Solid bars indicate the baseline.}
     \label{fig:zero-shot}
    \vspace{-1em}
\end{figure}

\paragraph{Results:} 
We report the ARM scores in Figure~\ref{fig:zero-shot}. Averaged over all rules, our approach obtains a ARM of 0.566, while the naive all-chance baseline only achieves 0.506. The difference appears to be small, but we still consider it significant, because these languages do not actually require agreement for many grammatical features.
Tagalog and Buryat are the most distant languages that we test on (no Philippine and Mongolic language is present in our training data) and yet we observe our method being at par with the baseline and even outperforming in case of Tagalog.
Breton and Welsh, on the other hand, are an interesting test bed: Celtic languages are to some degree outliers among Indo-European languages~\cite{borsley2005syntax}, and we suspect that as a result the parser performs generally worse. Despite that, our approach has an ARM of 0.730 for Welsh gender agreement, as opposed to the mere 0.615 that the baseline achieves.
 
\section{Related Work}
\citet{bender-etal-2014-learning} use interlinear glossed text (IGT) to extract lexical entities and morphological rules for an endangered language. 
They experiment with different systems which individually extract lemmas, lexical rules, word order and the case system, some of which use hand-specified rules. 
\citet{howell2017inferring} extend this to work to predict case system on additional languages. \citet{zamaraeva-2016-inferring} also infer morphotactics from IGT using $k$-means clustering. 
To the best of our knowledge, our work is the first to propose a framework to extract first-pass grammatical agreement rules directly from raw text
in a statistically-informed objective way.
A parallel line of work \cite{hellan-2010-descriptive} extracts  a \textit{construction profile} of a language by having templates that define how sentences are constructed.

\section{Future Work}


While we have demonstrated that our approach is effective in extracting a first-pass set of agreement rules directly from raw text, it focuses only on agreement between a pair of words and hence might fail to capture more complex phenomena that require broader context or operate at the phrase level.
Consider this simple English example: ``John and Mary love their dog''. Under both UD and SUD formalisms, the coordinating conjunction ``and" is a dependent, hence the verb will not agree with either of the (singular) nouns (``John" or ``Mary").  Also, deciding agreement based on only POS tags is insufficient to capture \emph{all} phenomena that may influence agreement for e.g. mass nouns such as `rice' do not follow the standard number agreement rules in English. We leave a more expressive model and evaluation on more languages as future work. We also plan to expand our methodology for extracting grammar rules from raw text to other aspects of morphosyntax, such as argument structure and word order phenomena.


\section*{Acknowledgments}
The authors are grateful to the anonymous reviewers who took the time to provide many interesting comments that made the paper significantly better, and to Josep Quer, Ekaterina Vylomova and Maria Ryskina,  for participating in the human annotation experiments. This work is sponsored by the DARPA grant FA8750-18-2-0018 and by the National Science Foundation under grant 1761548.

\bibliography{anthology,emnlp2020}
\bibliographystyle{acl_natbib}

\clearpage
\pagebreak
\appendix
\section{Appendix}

\subsection{Decision Tree Hyperparameters}
We perform a grid search over the following hyperparameters of the decision tree: 

\begin{itemize}
    \item \texttt{criterion} $=$ [gini, entropy]
    \item \texttt{max depth} $=$ [6,15]
    \item \texttt{min impurity decrease} $= 1e^{-3}$ 
\end{itemize}
The best parameters are selected based on the validation set performance. For some treebanks which have no validation set we use the default cross-validation provided by \texttt{sklearn} \cite{sklearn_api}.  Average model runtime for a treebanks is 5-10mins depending on the size of the treebank. 

\subsection{Dataset Statistics}

For the true low-resource experiments, the dataset details are in Table \ref{tab:realdata}.
\begin{table}[h]
\small
 \begin{center}{
 \begin{tabular}{l|l}
 \textbf{\textsc{Language}} & \textbf{\textsc{Train / Test}}\\
 \toprule
Breton-KEB & 30000 / 888  \\
Buryat-BXR & 10000  / 908  \\
Faroese-OFT & 50000/ 1208  \\
Tagalog-TRG & 30000  / 55   \\
Welsh-CCG & 30000 / 956

 \end{tabular}
 }
 \caption{ Dataset statistics.  Training data is obtained by parsing the Liepzig corpora \cite{goldhahn2012building} and test data is obtained from the respective treebank. Each cell denotes the number of sentences in train/test.}
 \label{tab:realdata}
 \end{center}

 \end{table}

\subsection{Evaluation}
\subsection{Annotation Interface for Expert Evaluation}
\begin{figure*}
\centering
\includegraphics[width=\textwidth]{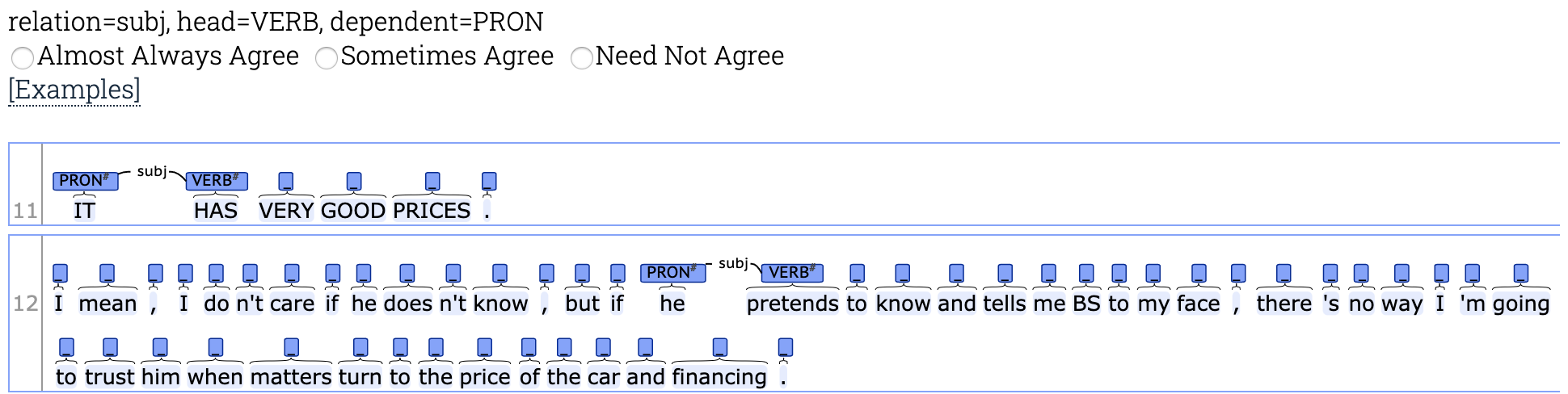}
  \caption{Annotation interface for evaluating \texttt{Gender} agreement in Catalan.}%
\label{fig:ling_annotation}%
\end{figure*}
In Figure \ref{fig:ling_annotation}, we show the annotation interface used for verifying \texttt{Gender} agreement rules in Catalan. For each triple, we display 10 randomly selected examples from the training portion of the treebank. 


\subsection{Low-resource Experiment Results}
For the simulation experiments, the dataset details are in Table \ref{tab:data}.

\subsubsection{Udify \cite{kondratyuk-straka-2019-75} Model Details}
We used the \texttt{Udify} model for automatically annotating the raw text with part-of-speech (POS), dependency links and morphological features. For each of the simulation experiment we report the \texttt{udify} parsing performance on the test data in Table \ref{tab:parsing}. We used the same hyperparameters for training with a related languages as specified by the authors.\footnote{\url{https://github.com/Hyperparticle/udify}}. In the configuration file, we only change the parameters \texttt{warmup steps}$=100$ and \texttt{start-step}$=100$, as recommended by the authors for low-resource languages.

\subsubsection{Results and Discussion}
For each language and feature, we plot the ARM score with and without transfer learning in Figure \ref{fig:gender}-\ref{fig:case}. Similar to our findings for \texttt{Gender} in Figure 5, we find that cross-lingual transfer leads to a better score across all languages in the zero-shot setting. As we increase the number of gold-standard sentences, the quality of extracted rules improve. Although, for Belarusian we observe the opposite trend for \texttt{Person} agreement. On closer inspection we find that it is because person applies only to non-past finite verb forms (VERB and AUX) as an inflectional feature and to pronouns (PRON) as a lexical feature which means that in many cases person is not explicitly marked, even though it implicitly exists \footnote{\url{https://universaldependencies.org/be/}}. 

  \begin{table*}[h]
 \begin{center} {
 \begin{tabular}{l|l|l}
 \textbf{\textsc{Language}} & \textbf{\textsc{Train/Dev/Test}} & \textbf{\textsc{Transfer Language}} \\
 \toprule
Spanish-GSD & 14187 / 1400/ 426 & Portuguese-Bosque \\
Greek-GDT & 1662 / 403 / 456 & Ancient Greek-PROIEL \\
Belarusian-HSE & 319 / 65/ 253 & Ukrainian-IU \\
Lithuanian-ALKSNIS & 2341 / 617 / 684  & Latvian-LVTB \\

 \end{tabular}
 }
 \caption{ Dataset statistics.  Train/Dev/Test denote the number of sentences in the respective treebank used for the target language.}
 \label{tab:data}
 \end{center}

 \end{table*}

\begin{table*}[h]
\small
 \begin{center}{
 \begin{tabular}{l|l|l|l}
 \textbf{\textsc{Language}}  & \textbf{\textsc{\#training}} & \multicolumn{2}{c}{\textbf{\textsc{setting}}} \\
 & &  \textbf{\textsc{w/o Transfer}}  & \textbf{\textsc{+Transfer}}\\ 
 \toprule
Greek & 0 & - & upos:0.661, ufeats:0.392, uas:0.632, las :0.465 \\
& 50 & upos:0.507, ufeats:0.330, uas:0.309, las:0.203 & upos:0.877, ufeats:0.631, uas:0.724, las:0.653\\
& 100 & upos:0.915, ufeats:0.664, uas: 0.755, las: 0.691& upos: 0.906, ufeats: 0.719, uas: 0.758, las: 0.703\\
& 500 & upos: 0.970, ufeats: 0.891, uas: 0.891, las: 0.866 & upos: 0.954, ufeats: 0.860, uas: 0.849, las: 0.817 \\
\midrule
Spanish & 0 & - & upos: 0.922, ufeats: 0.764, uas: 0.855, las: 0.776 \\
& 50 & upos: 0.529, ufeats: 0.463, uas: 0.289, las: 0.152& upos: 0.913, ufeats: 0.792, , uas: 0.844, las: 0.767\\
& 100 & upos: 0.920, ufeats: 0.832, uas: 0.755, las: 0.690 & upos: 0.916, ufeats: 0.840, uas: 0.849, las: 0.784 \\
& 500 & upos: 0.952, ufeats: 0.919, uas: 0.860, las: 0.820 & upos: 0.949, ufeats: 0.889, uas: 0.859, las: 0.822 \\
\midrule
Belarusian & 0 & - & 
upos: 0.941, ufeats: 0.520, uas: 0.863, las: 0.797 \\
& 50 & upos: 0.570, ufeats: 0.323, uas: 0.217, las: 0.141 & upos: 0.952, ufeats: 0.726, uas: 0.763, las: 0.727\\
& 100 & upos: 0.919, ufeats: 0.446, uas: 0.521, las: 0.482 & upos: 0.961, ufeats: 0.777, uas: 0.854, las: 0.800 \\
\midrule
Lithuanian & 0 & - & upos: 0.869, ufeats: 0.528, uas: 0.752, las: 0.610\\
& 50 & upos: 0.566, ufeats: 0.371, uas: 0.346, las: 0.211 & upos: 0.874, ufeats: 0.5841, uas: 0.757, las: 0.623 \\
& 100 & upos: 0.813, ufeats: 0.453, uas: 0.551, las: 0.421 &upos: 0.883, ufeats: 0.637, uas: 0.761, las: 0.659\\
& 500 & upos: 0.925, ufeats: 0.744, uas: 0.757, las: 0.697& upos: 0.912, ufeats: 0.747, uas: 0.779, las: 0.714\\

 \end{tabular}
 }
 \caption{\texttt{udify} model performance on the test data for each low-resource setting. The scores are averaged across five runs of each setting.}
 \label{tab:parsing}
 \end{center}

\end{table*}
 \begin{figure*}[h]
     \centering
     \includegraphics[width=\columnwidth]{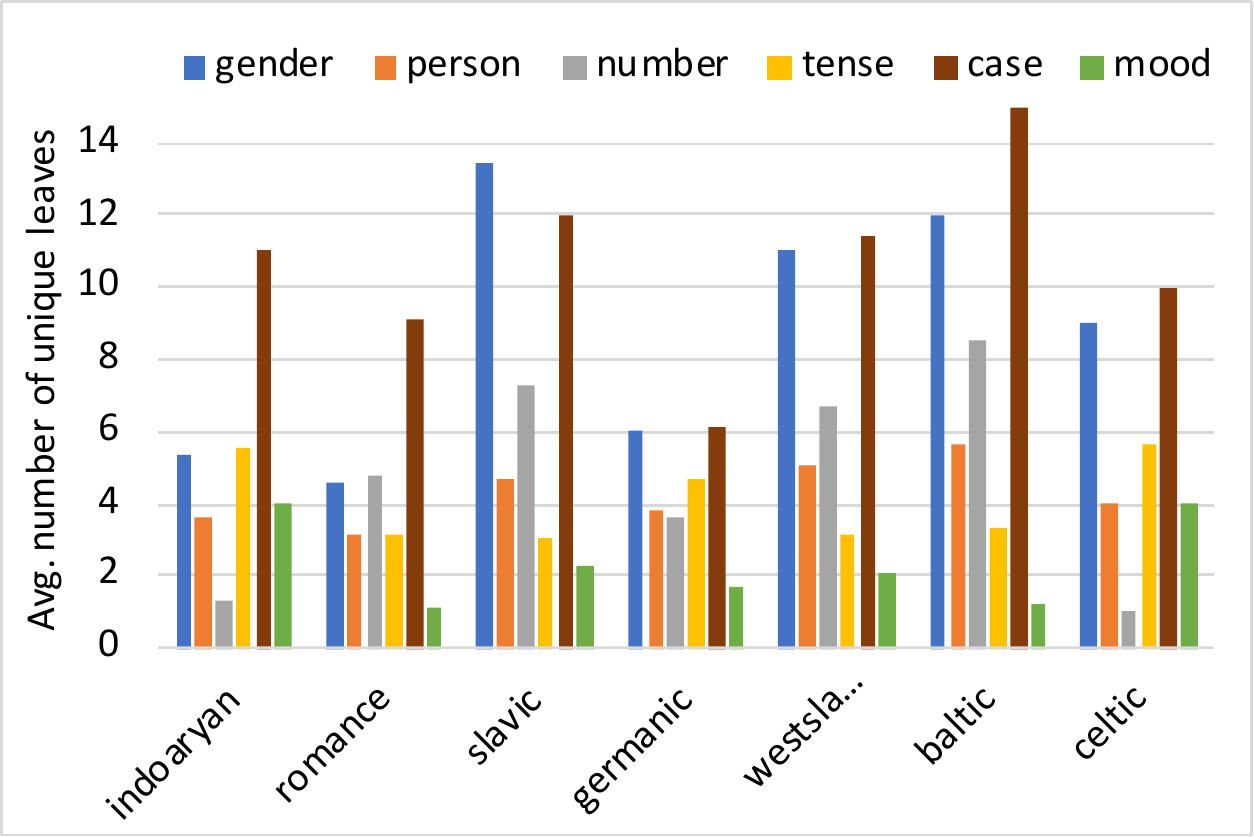}
     \caption{(Avg.) number of leaves for each feature grouped by language family.}
     \label{fig:leaves}
 \end{figure*}
\subsection{Experiments with Gold-Standard Data}
 We present the ARM scores for all treebanks and features in Tables \ref{tab:all}-\ref{tab:all_9}. We also report the validation results in the same tables for our best setting which uses the \textit{Statistical Threshold}. 
 In Section 2.2, we proposed using two types of thresholds for retaining the high probability agreement rules. In order to compare which threshold is the best for all treebanks, we manually inspect some of the learnt decision trees. We find that for the trees  learnt from the \textit{hard threshold} often over-fit on the training data causing to produce leaves with very few examples. In Figure \ref{fig:sud} we compare the trees constructed for number agreement with the two thresholds for Marathi. One reason why \textit{Statistical-Threshold} performs better for low-resource languages is because there are more leaves with fewer samples overall causing the \textit{Hard Threshold} to have more false positives. Whereas the \textit{Statistical Threshold} uses  \emph{effect size} with the significance test which takes into account the sample size within a leaf leading to better leaves. Therefore, we choose to use \textit{Statistical-Threshold} for all our simulation experiments.
 
 In Figure \ref{fig:leaves}, we report that (avg.) number of leaves in the decision trees grouped by language family. Overall, \texttt{Gender} and \texttt{Case} tend to have more complex trees. For \texttt{Case}, it is probably because languages have more number of cases making it harder for the decision tree to model them.

\begin{figure*}%
\centering
\subfigure[]{
\label{}%
\includegraphics[width=0.4\textwidth]{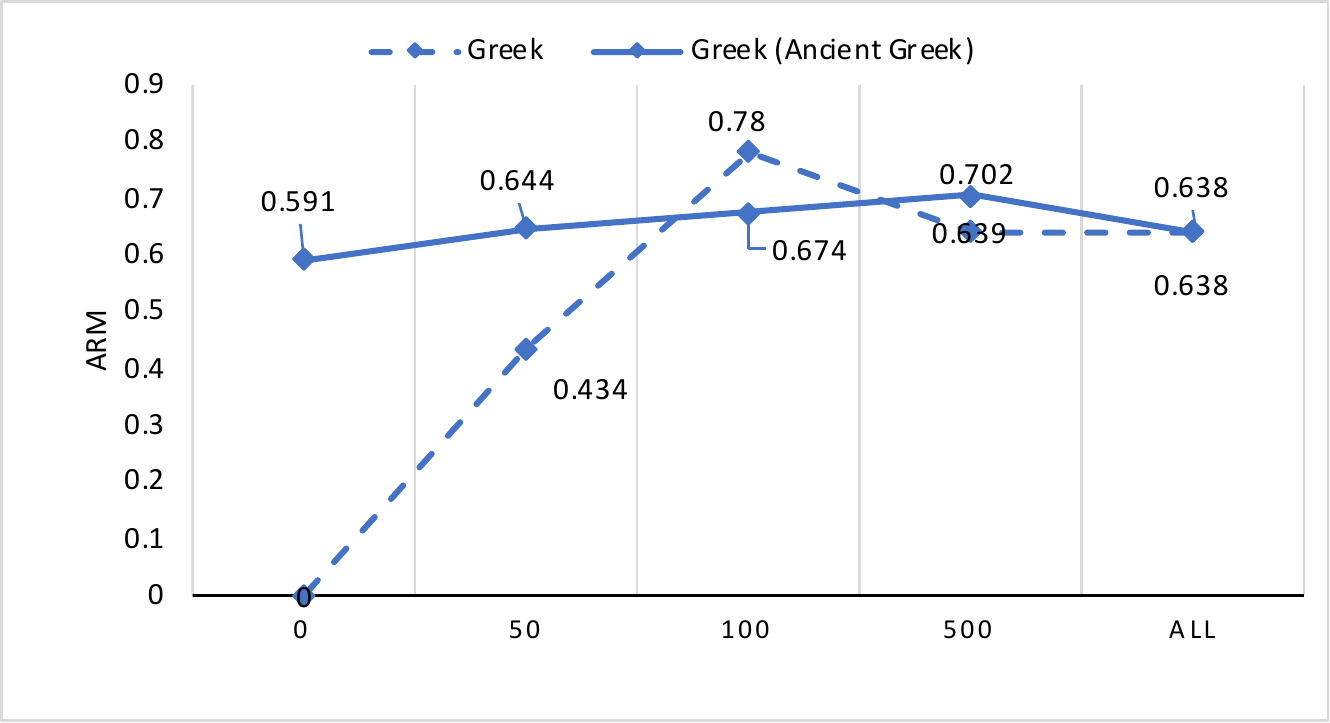}}
~
\subfigure[]{
\label{tree}%
\includegraphics[width=0.4\textwidth]{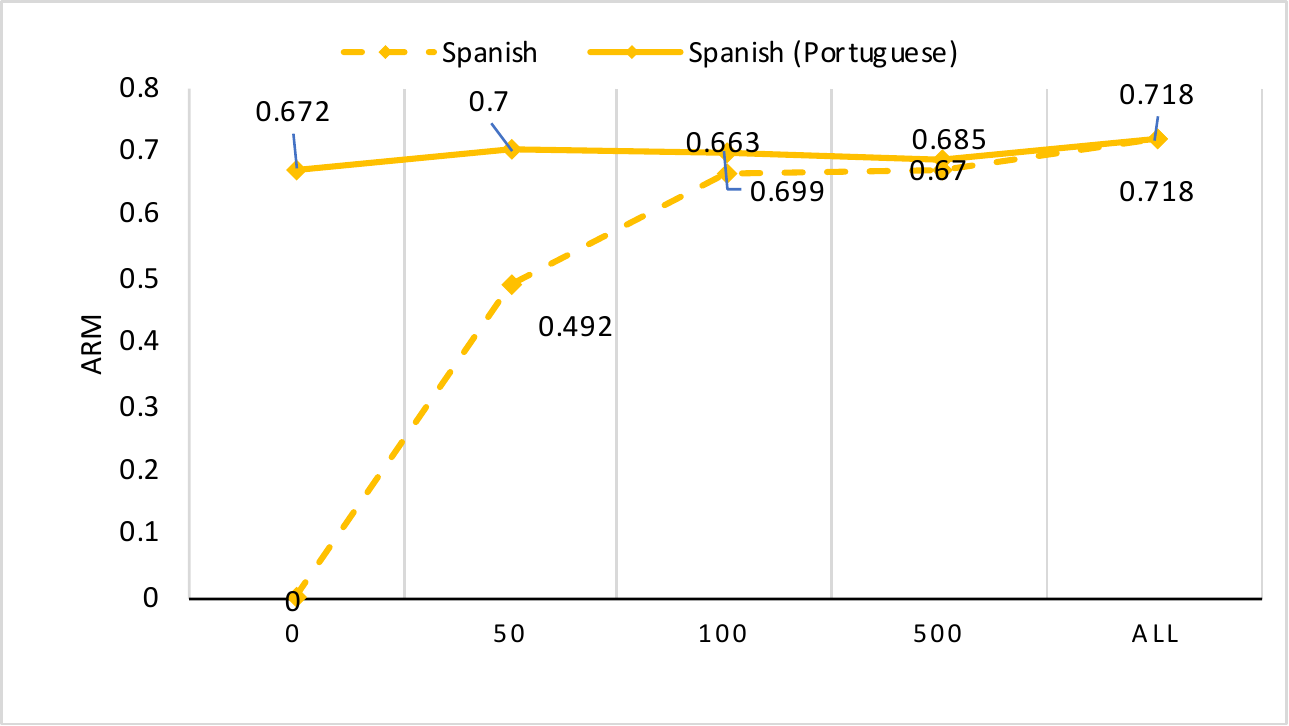}
}
~
\subfigure[]{
\label{}%
\includegraphics[width=0.4\textwidth]{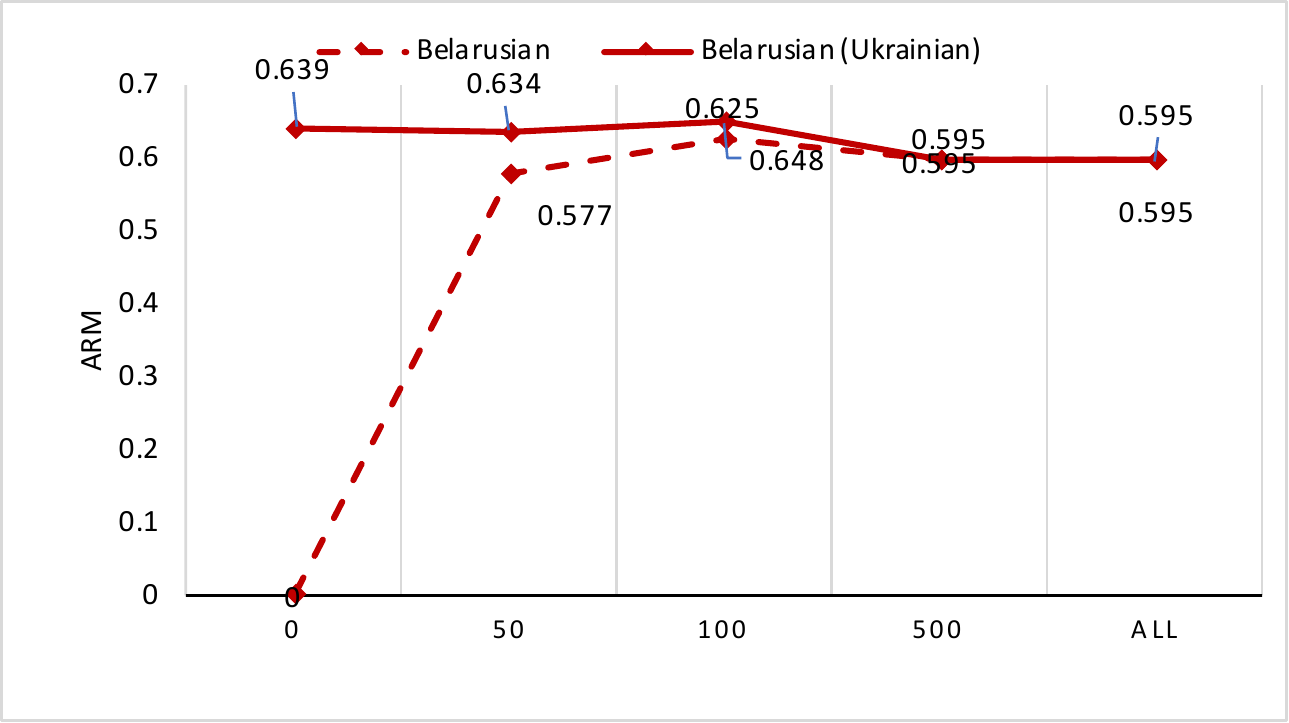}}
~
\subfigure[]{
\label{}%
\includegraphics[width=0.4\textwidth]{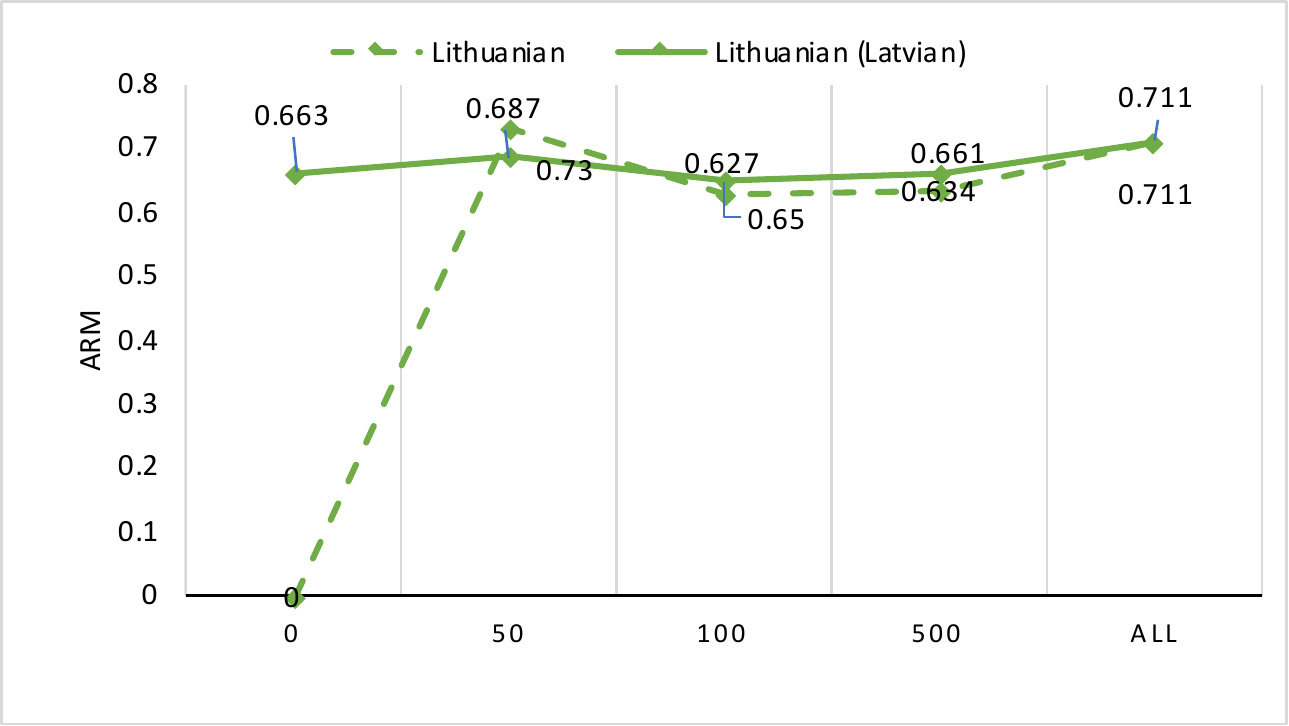}}
~
  \caption{Comparing the (avg.) ARM score for \texttt{Gender} agreement with and without cross-lingual transfer learning (transfer language in parenthesis). Note: the higher the ARM the better.  }%
\label{fig:gender}%
\end{figure*}
\begin{figure*}%
\centering
\subfigure[]{
\label{}%
\includegraphics[width=0.4\textwidth]{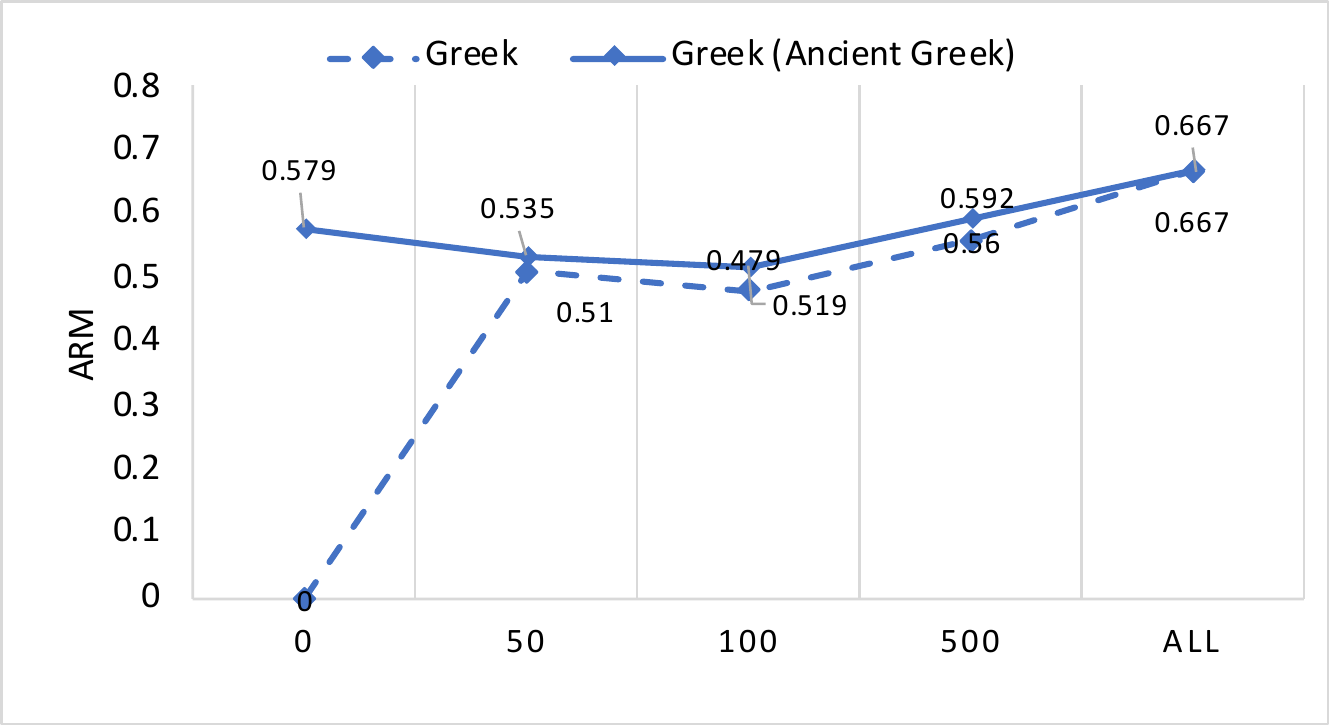}}
~
\subfigure[]{
\label{tree}%
\includegraphics[width=0.4\textwidth]{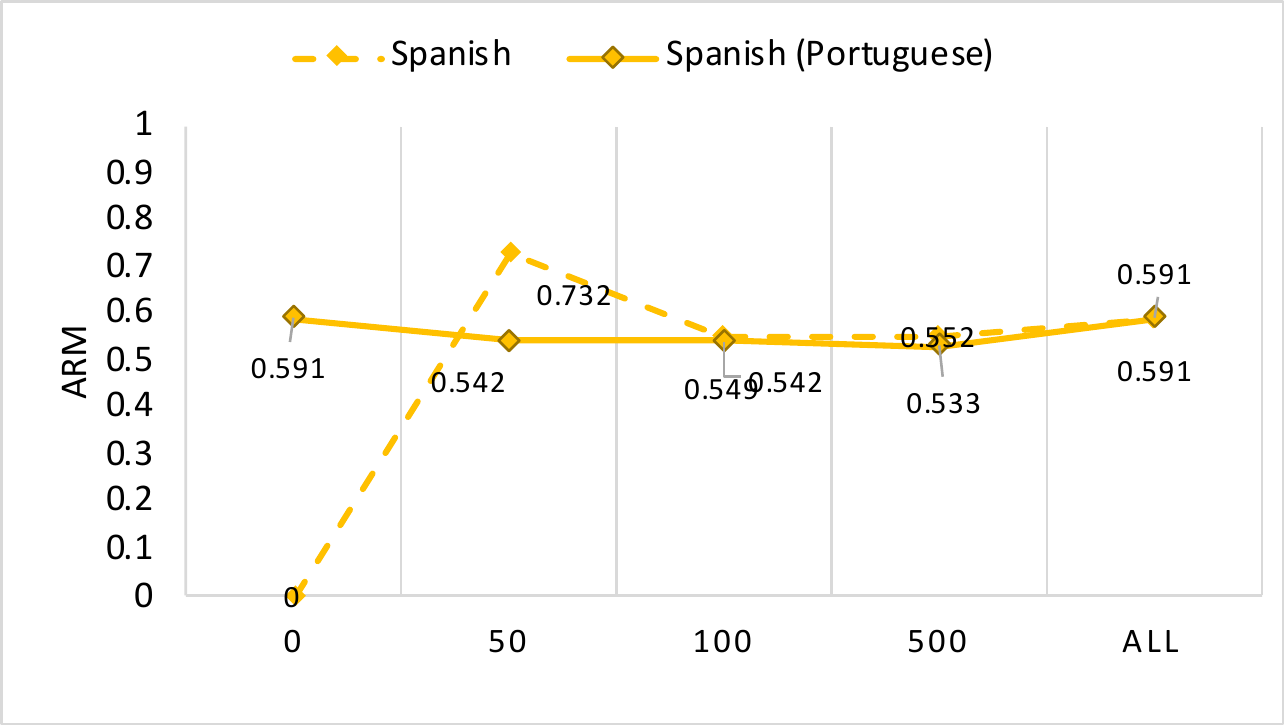}
}
~
\subfigure[]{
\label{}%
\includegraphics[width=0.4\textwidth]{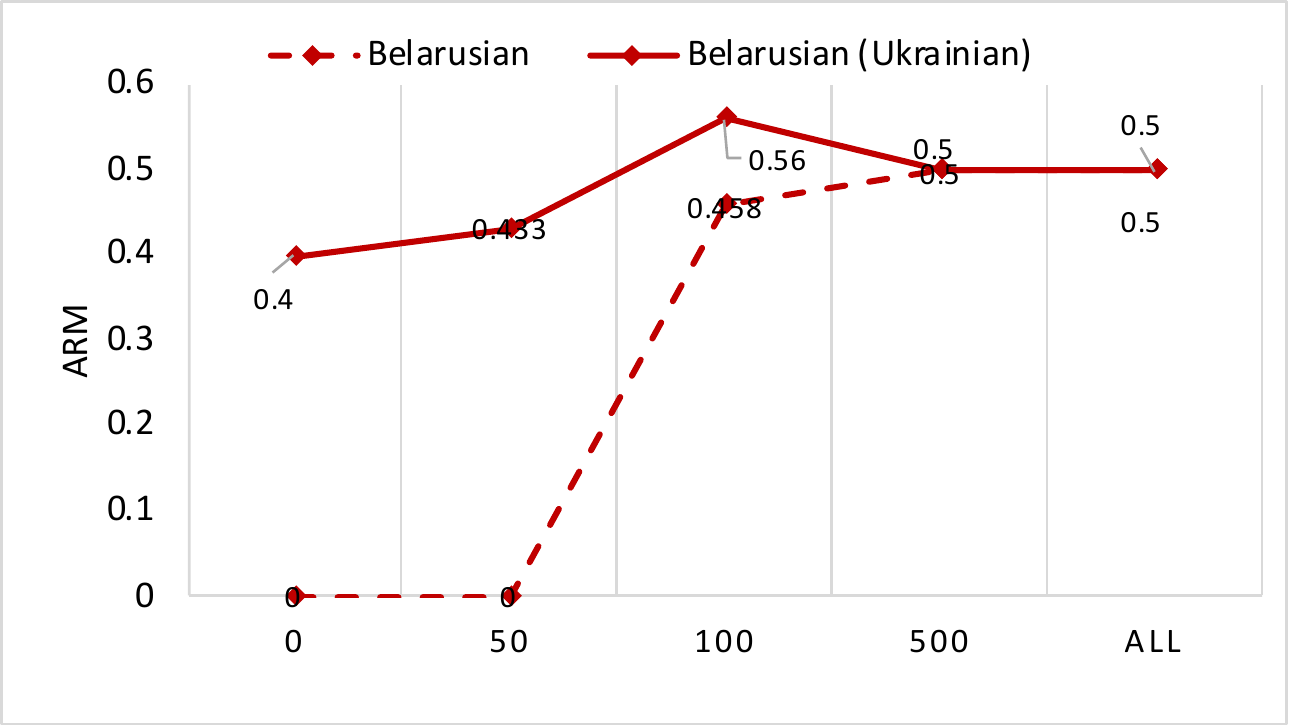}}
~
\subfigure[]{
\label{}%
\includegraphics[width=0.4\textwidth]{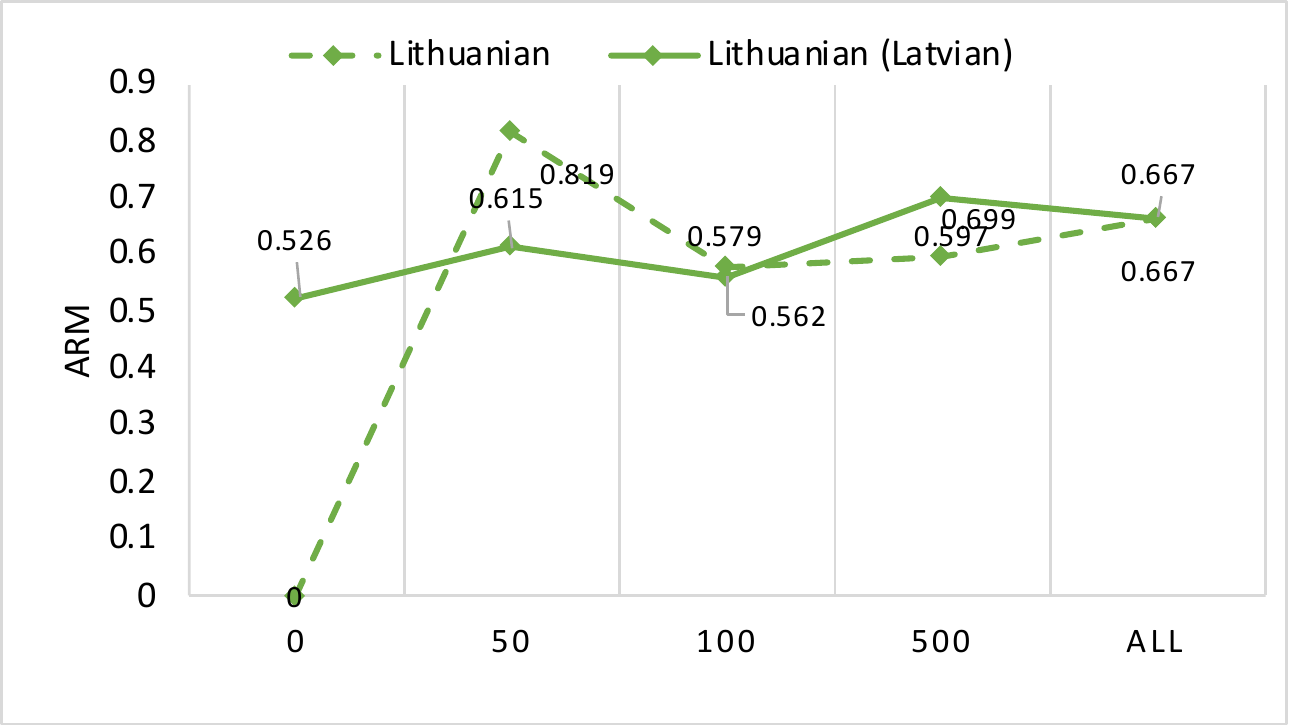}}
~
  \caption{Comparing the (avg.) ARM score for \texttt{Person} agreement with and without cross-lingual transfer learning (transfer language in parenthesis). Note: the higher the ARM the better.  }%
\label{fig:person}%
\end{figure*}
\begin{figure*}%
\centering
\subfigure[]{
\label{}%
\includegraphics[width=0.4\textwidth]{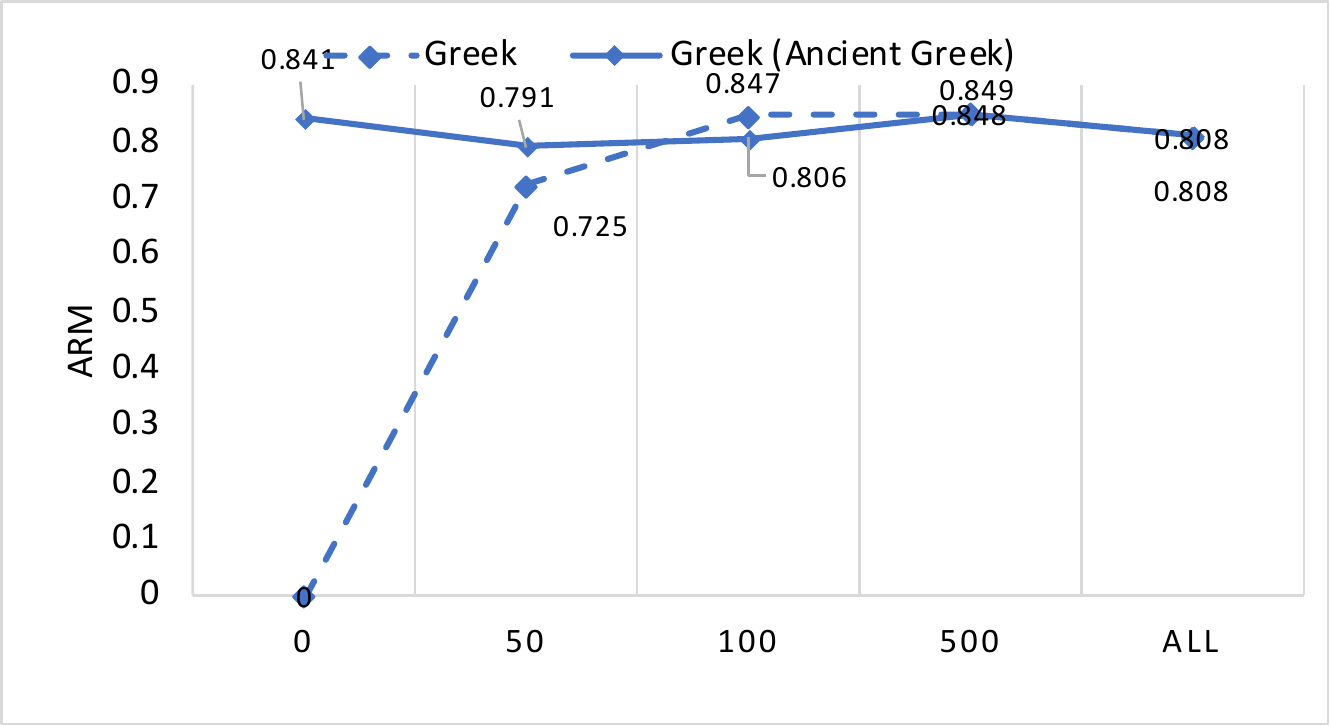}}
~
\subfigure[]{
\label{tree}%
\includegraphics[width=0.4\textwidth]{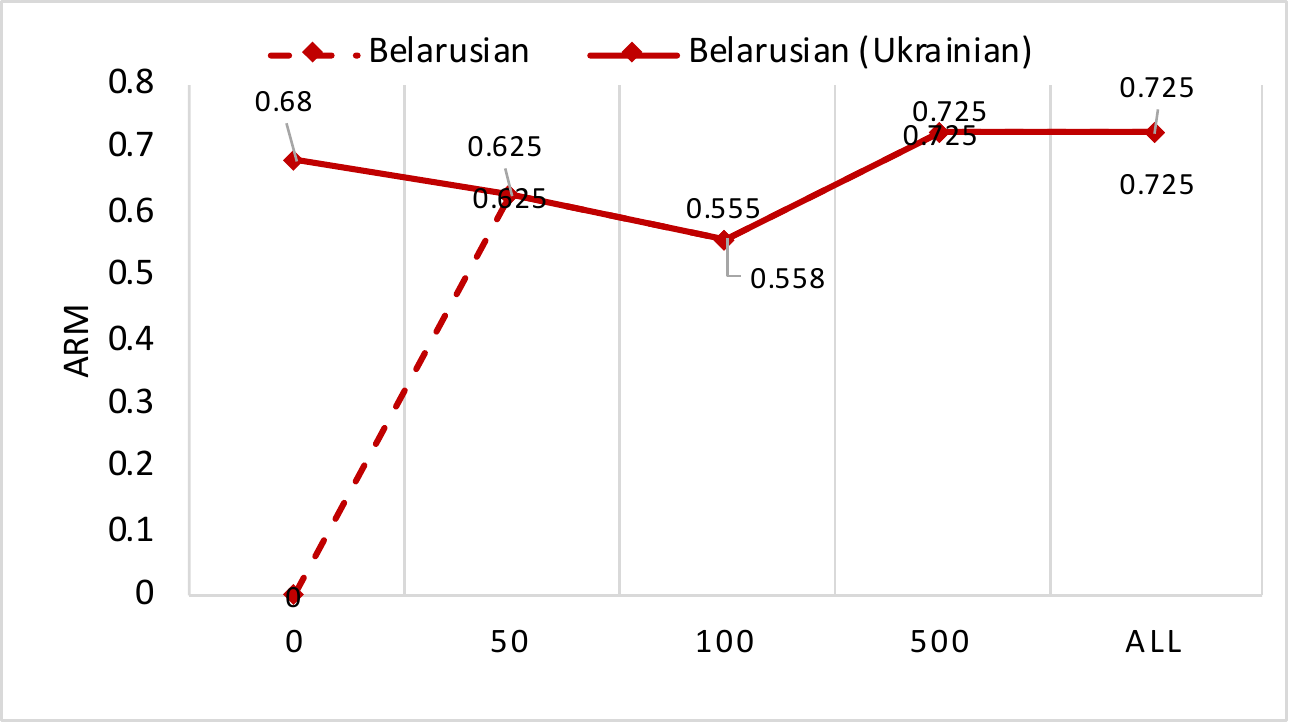}
}
~
\subfigure[]{
\label{}%
\includegraphics[width=0.4\textwidth]{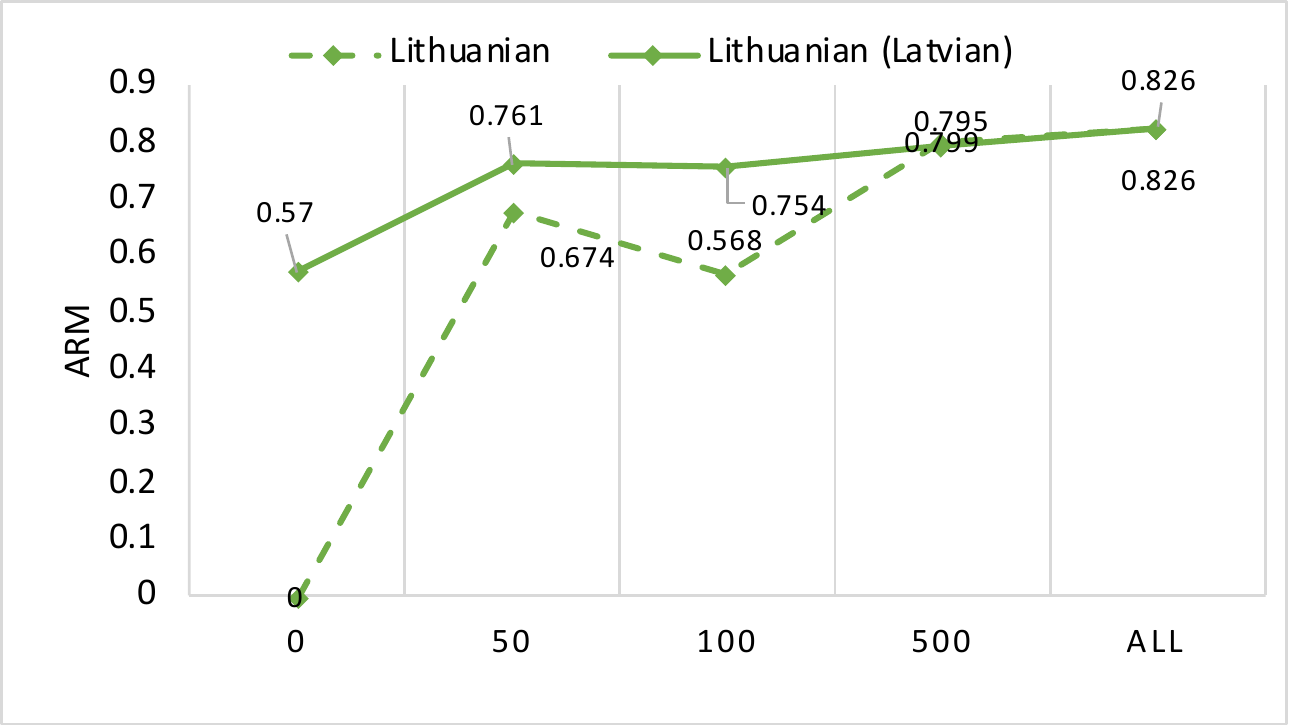}}

  \caption{Comparing the (avg.) ARM score for \texttt{Case} agreement with and without cross-lingual transfer learning (transfer language in parenthesis). Note: the higher the ARM the better. For Spanish, there was $<10$ data points with \texttt{Case} annotated hence we do not report results for it. }%
\label{fig:case}%
\end{figure*}

\begin{figure*}%
\centering
\subfigure[]{
\label{}%
\includegraphics[width=\textwidth]{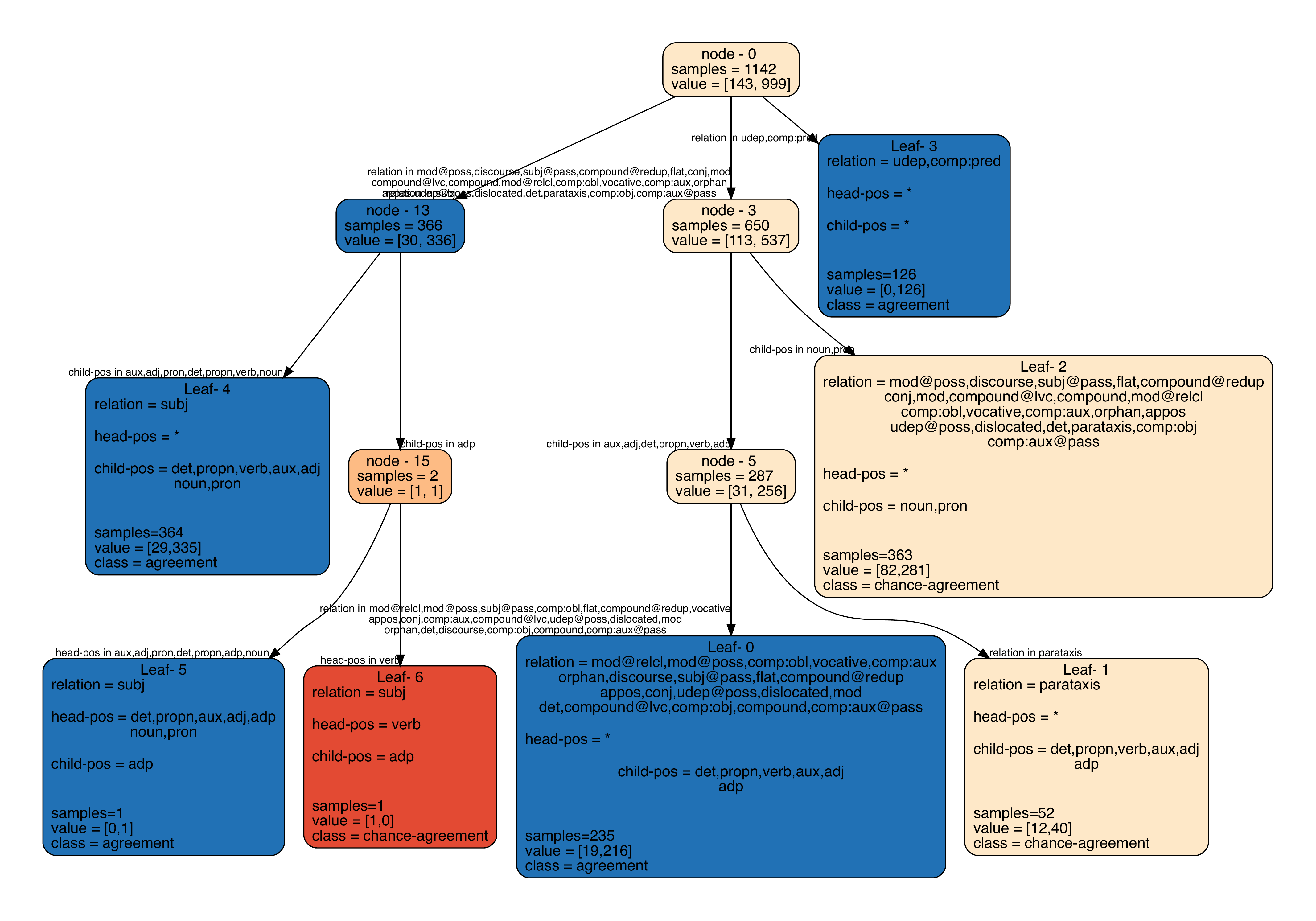}}
~
\subfigure[]{
\includegraphics[width=\textwidth]{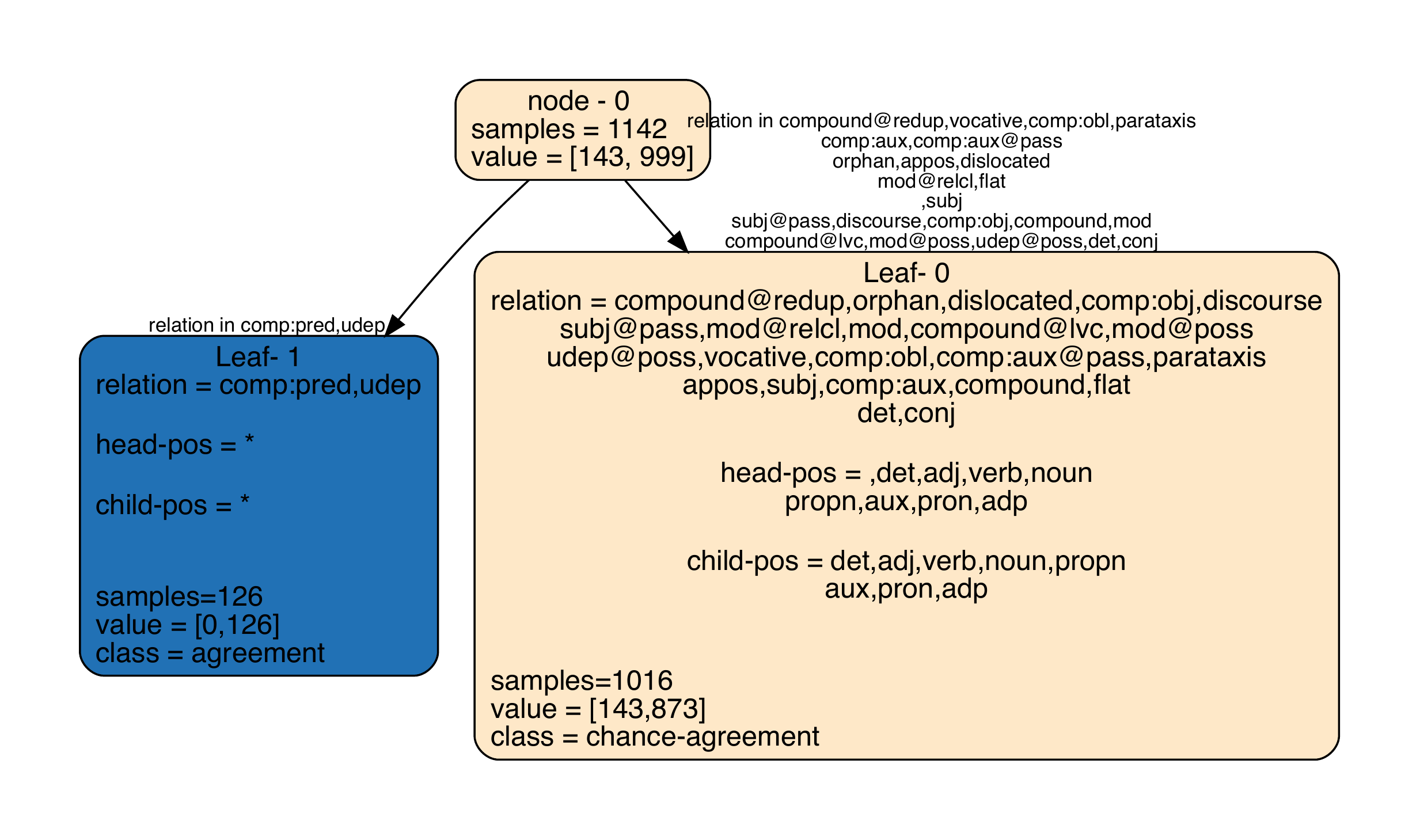}
}
~
  \caption{Comparing the learnt trees for \texttt{Number} agreement extracted using (a) \textit{Hard Threshold} and (b) \textit{Statistical Threshold}. \textit{Hard Threshold} overfits on the training data resulting in leaves with very few samples.  }%
\label{fig:sud}%
\end{figure*}

\subsection{SUD treebanks}

\begin{figure*}[h]%
\centering
\subfigure[]{
\label{}%
\includegraphics[width=0.4\textwidth]{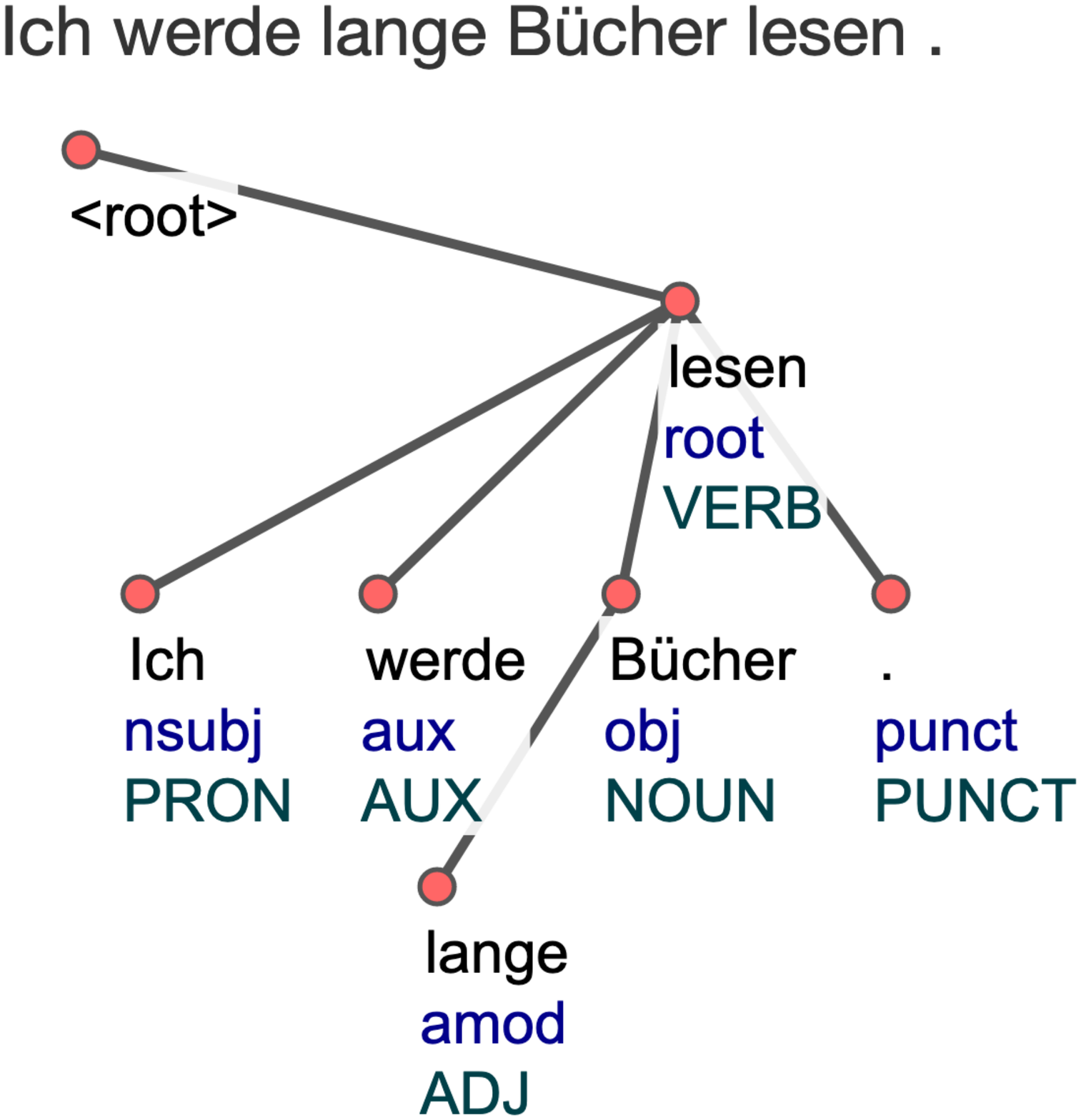}}
~
\subfigure[]{
\label{}%
\includegraphics[width=0.4\textwidth]{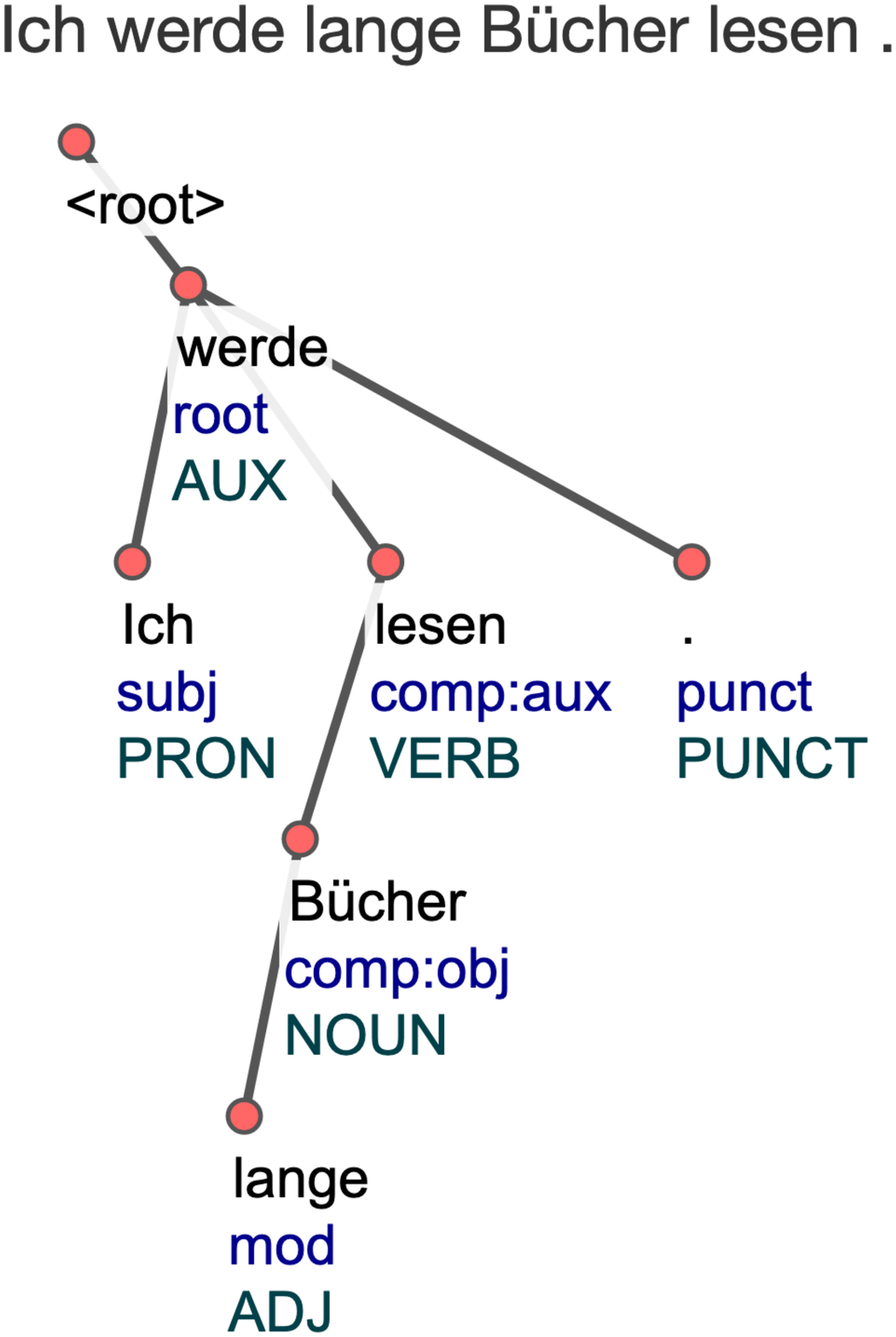}
}
~
  \caption{Comparing the UD (a) tree with the SUD (b) tree for the German sentence ``Ich werde lange B\"{u}cher lesen.".  }%
\label{fig:ud_sud_example}%
\end{figure*}

Figure~\ref{fig:ud_sud_example} presents a comparison of UD and SUD-style trees for the German sentence, ``Ich werde lange B\"{u}cher lesen.". The SUD tree has the function word `werde' as the syntactic head to the content word `lesen'.

\begin{table*}[h]
\small
 \begin{center}{
 \begin{tabular}{l|l|l|l|l|l}
 \textbf{\textsc{treebank}} &  \textbf{\textsc{feature}} & \textbf{\textsc{Statistical}} & \textbf{\textsc{Hard}} &
 \textbf{\textsc{Baseline}} &
\textbf{\textsc{Dev}} \\

 \toprule
ru-gsd & Gender & 0.678 & - & 0.51 & 0.623  \\ 
ru-gsd & Person & 0.125 & 0.875 & 0.125 & 0.286  \\ 
ru-gsd & Number & 0.628 & 0.512 & 0.384 & 0.62  \\ 
ru-gsd & Tense & 0.667 & 0.667 & 0.667 & 0.571  \\ 
ru-gsd & Mood & 0.0 & 1.0 & 0.0 & 0.1  \\ 
ru-gsd & Case & 0.649 & 0.614 & 0.395 & 0.537  \\ 
id-gsd & Number & 0.047 & 0.961 & 0.047 & 0.045  \\ 
it-isdt & Gender & 0.816 & 0.816 & 0.289 & 0.738  \\ 
it-isdt & Person & 0.304 & 0.87 & 0.304 & 0.619  \\ 
it-isdt & Number & 0.615 & 0.603 & 0.41 & 0.588  \\ 
it-isdt & Tense & 0.765 & 0.765 & 0.647 & 0.611  \\ 
it-isdt & Mood & 0.25 & 0.75 & 0.25 & 0.273  \\ 
la-proiel & Gender & 0.538 & 0.568 & 0.636 & 0.496  \\ 
la-proiel & Person & 0.56 & 0.6 & 0.54 & 0.653  \\ 
la-proiel & Number & 0.648 & 0.574 & 0.452 & 0.553  \\ 
la-proiel & Tense & 0.818 & 0.879 & 0.879 & 0.824  \\ 
la-proiel & Mood & 0.6 & 0.52 & 0.44 & 0.667  \\ 
la-proiel & Case & 0.759 & 0.782 & 0.466 & 0.691  \\ 
ro-nonstandard & Gender & 0.64 & 0.57 & 0.407 & 0.75  \\ 
ro-nonstandard & Person & 0.636 & 0.606 & 0.606 & 0.683  \\ 
ro-nonstandard & Number & 0.626 & 0.626 & 0.586 & 0.693  \\ 
ro-nonstandard & Tense & 0.452 & 0.839 & 0.645 & 0.467  \\ 
ro-nonstandard & Mood & 0.676 & 0.765 & 0.676 & 0.4  \\ 
ro-nonstandard & Case & 0.694 & 0.702 & 0.636 & 0.704  \\ 
he-htb & Gender & 0.747 & 0.747 & 0.663 & 0.629  \\ 
he-htb & Person & 0.737 & 0.789 & 0.789 & 0.769  \\ 
he-htb & Number & 0.585 & 0.585 & 0.415 & 0.505  \\ 
he-htb & Tense & 0.3 & 0.3 & 0.1 & 0.545  \\ 
he-htb & Case & 0.5 & 0.5 & 0.0 & 0.5  \\ 
no-bokmaal & Gender & 0.477 & 0.568 & 0.545 & 0.675  \\ 
no-bokmaal & Person & 1.0 & 1.0 & 0.5 & 1.0  \\ 
no-bokmaal & Number & 0.655 & 0.673 & 0.364 & 0.733  \\ 
no-bokmaal & Tense & 0.55 & 0.55 & 0.55 & 0.55  \\ 
no-bokmaal & Mood & 0.0 & 1.0 & 0.0 & 0.1  \\ 
no-bokmaal & Case & 0.0 & 0.333 & 0.0 & 0.0  \\ 
no-nynorsk & Gender & 0.464 & 0.536 & 0.536 & 0.514  \\ 
no-nynorsk & Person & 0.0 & 0.0 & 0.0 & 0.667  \\ 
no-nynorsk & Number & 0.702 & 0.702 & 0.511 & 0.596  \\ 
no-nynorsk & Tense & 0.368 & 0.368 & 0.684 & 0.429  \\ 
no-nynorsk & Mood & 0.0 & 1.0 & 0.0 & 0.048  \\ 
fi-tdt & Person & 0.387 & 0.677 & 0.677 & 0.607  \\ 
fi-tdt & Number & 0.502 & 0.493 & 0.511 & 0.559  \\ 
fi-tdt & Tense & 0.474 & 0.368 & 0.474 & 0.5  \\ 
fi-tdt & Mood & 0.75 & 0.75 & 0.75 & 0.471  \\ 
fi-tdt & Case & 0.786 & 0.828 & 0.821 & 0.781  \\ 
pl-lfg & Gender & 0.646 & 0.646 & 0.463 & 0.641  \\ 
pl-lfg & Person & 0.688 & 0.688 & 0.562 & 0.714  \\ 
pl-lfg & Number & 0.691 & 0.68 & 0.412 & 0.624  \\ 
pl-lfg & Tense & 0.556 & 0.667 & 0.667 & 0.6  \\ 
pl-lfg & Mood & 0.333 & 0.667 & 0.333 & 0.4  \\ 
pl-lfg & Case & 0.744 & 0.667 & 0.41 & 0.617  \\ 
grc-perseus & Gender & 0.62 & 0.718 & 0.563 & 0.699  \\ 
grc-perseus & Person & 0.8 & 0.8 & 0.7 & 0.636  \\ 
grc-perseus & Number & 0.531 & 0.63 & 0.605 & 0.537  \\ 
grc-perseus & Tense & 0.889 & 1.0 & 1.0 & 0.778  \\ 
grc-perseus & Mood & 0.833 & 0.833 & 0.667 & 0.429  \\ 
grc-perseus & Case & 0.708 & 0.792 & 0.556 & 0.712  \\ 
fi-ftb & Person & 0.56 & 0.76 & 0.6 & 0.63  \\ 
fi-ftb & Number & 0.524 & 0.441 & 0.524 & 0.54  \\ 
fi-ftb & Tense & 0.846 & 0.769 & 0.308 & 0.538  \\ 
fi-ftb & Mood & 0.429 & 0.5 & 0.429 & 0.529  \\ 
fi-ftb & Case & 0.724 & 0.848 & 0.781 & 0.748  \\ 
wo-wtb & Gender & 0.5 & 0.5 & 0.5 & 0.0  \\ 
wo-wtb & Person & 0.55 & 0.45 & 0.4 & 0.609  \\ 
wo-wtb & Number & 0.486 & 0.6 & 0.6 & 0.632  \\ 
wo-wtb & Tense & 0.5 & 0.625 & 0.375 & 0.625  \\ 
wo-wtb & Mood & 0.143 & 0.143 & 0.143 & 0.364  \\ 
en-partut & Person & 0.5 & 0.5 & 0.417 & 0.857  \\ 
en-partut & Number & 0.559 & 0.559 & 0.441 & 0.676  \\ 
en-partut & Tense & 0.667 & 0.733 & 0.667 & 0.583  \\ 
en-partut & Mood & 0.091 & 0.818 & 0.091 & 0.1  \\ 

\end{tabular}
 }
 \caption{Comparing the ARM scores for SUD treebanks across both Statistical and Hard thresholding.}
 \label{tab:all}
 \end{center}

 \end{table*}

\begin{table*}[h]
\small
 \begin{center}{
 \begin{tabular}{l|l|l|l|l|l}
 \textbf{\textsc{treebank}} &  \textbf{\textsc{feature}} & \textbf{\textsc{Statistical}} & \textbf{\textsc{Hard}} &
 \textbf{\textsc{Baseline}} &
\textbf{\textsc{Dev}} \\
 \toprule
 fr-ftb & Gender & 0.631 & 0.631 & 0.477 & 0.621  \\ 
fr-ftb & Person & 0.14 & 0.86 & 0.14 & 0.171  \\ 
fr-ftb & Number & 0.635 & 0.635 & 0.502 & 0.634  \\ 
fr-ftb & Tense & 0.714 & 0.857 & 0.857 & 0.833  \\ 
fr-ftb & Mood & 0.409 & 0.591 & 0.409 & 0.6  \\ 
lv-lvtb & Gender & 0.727 & 0.734 & 0.461 & 0.677  \\ 
lv-lvtb & Person & 0.5 & 0.632 & 0.579 & 0.583  \\ 
lv-lvtb & Number & 0.688 & 0.688 & 0.429 & 0.706  \\ 
lv-lvtb & Tense & 0.667 & 0.815 & 0.889 & 0.741  \\ 
lv-lvtb & Mood & 0.476 & 0.619 & 0.476 & 0.333  \\ 
lv-lvtb & Case & 0.719 & 0.734 & 0.489 & 0.772  \\ 
ro-rrt & Gender & 0.583 & 0.583 & 0.51 & 0.591  \\ 
ro-rrt & Person & 0.327 & 0.755 & 0.347 & 0.304  \\ 
ro-rrt & Number & 0.535 & 0.585 & 0.528 & 0.56  \\ 
ro-rrt & Tense & 0.421 & 0.684 & 0.789 & 0.526  \\ 
ro-rrt & Mood & 0.931 & 1.0 & 0.448 & 0.867  \\ 
ro-rrt & Case & 0.862 & 0.788 & 0.588 & 0.854  \\ 
it-vit & Gender & 0.672 & 0.672 & 0.375 & 0.678  \\ 
it-vit & Person & 0.625 & 0.625 & 0.792 & 0.667  \\ 
it-vit & Number & 0.712 & 0.728 & 0.528 & 0.61  \\ 
it-vit & Tense & 0.773 & 0.955 & 0.955 & 0.75  \\ 
it-vit & Mood & 0.4 & 0.6 & 0.4 & 0.231  \\
fr-partut & Gender & 0.579 & 0.632 & 0.421 & 0.615  \\ 
fr-partut & Person & 0.818 & 0.727 & 0.273 & 0.75  \\ 
fr-partut & Number & 0.771 & 0.542 & 0.292 & 0.542  \\ 
fr-partut & Tense & 0.857 & 0.857 & 0.714 & 0.6  \\ 
fr-partut & Mood & 0.333 & 0.667 & 0.333 & 0.167  \\ 
en-ewt & Person & 0.812 & 0.812 & 0.25 & 0.85  \\ 
en-ewt & Number & 0.357 & 0.643 & 0.357 & 0.304  \\ 
en-ewt & Tense & 0.591 & 0.773 & 0.773 & 0.593  \\ 
en-ewt & Mood & 0.4 & 0.733 & 0.4 & 0.333  \\ 
ru-syntagrus & Gender & 0.697 & 0.747 & 0.624 & 0.673  \\ 
ru-syntagrus & Person & 0.625 & 0.667 & 0.667 & 0.72  \\ 
ru-syntagrus & Number & 0.591 & 0.661 & 0.562 & 0.576  \\ 
ru-syntagrus & Tense & 0.727 & 0.818 & 0.818 & 0.667  \\ 
ru-syntagrus & Mood & 0.4 & 0.8 & 0.44 & 0.407  \\ 
ru-syntagrus & Case & 0.649 & 0.707 & 0.575 & 0.681  \\ 
sv-talbanken & Gender & 0.719 & 0.719 & 0.438 & 0.643  \\ 
sv-talbanken & Number & 0.659 & 0.634 & 0.463 & 0.571  \\ 
sv-talbanken & Tense & 0.559 & 0.588 & 0.5 & 0.607  \\ 
sv-talbanken & Mood & 0.048 & 0.952 & 0.048 & 0.056  \\ 
sv-talbanken & Case & 0.189 & 0.623 & 0.189 & 0.143  \\ 
olo-kkpp & Person & 0.286 & 0.571 & 0.286 & -  \\ 
olo-kkpp & Number & 0.667 & 0.692 & 0.667 & -  \\ 
olo-kkpp & Tense & 0.75 & 0.75 & 0.75 & -  \\ 
olo-kkpp & Mood & 0.0 & 0.75 & 0.0 & -  \\ 
olo-kkpp & Case & 0.7 & 0.7 & 0.7 & -  \\ 
cs-cac & Gender & 0.663 & 0.673 & 0.602 & 0.678  \\ 
cs-cac & Person & 0.562 & 0.562 & 0.5 & 0.583  \\ 
cs-cac & Number & 0.636 & 0.531 & 0.469 & 0.575  \\ 
cs-cac & Tense & 0.467 & 0.667 & 0.6 & 0.333  \\ 
cs-cac & Mood & 0.2 & 0.4 & 0.2 & 0.111  \\ 
cs-cac & Case & 0.81 & 0.84 & 0.46 & 0.833  \\ 
ur-udtb & Gender & 0.567 & 0.567 & 0.536 & 0.576  \\ 
ur-udtb & Person & 0.152 & 0.946 & 0.065 & 0.195  \\ 
ur-udtb & Number & 0.485 & 0.583 & 0.485 & 0.496  \\ 
ur-udtb & Tense & 0.333 & 0.5 & 0.5 & 0.667  \\ 
ur-udtb & Mood & 0.714 & 0.714 & 0.143 & 0.714  \\ 
ur-udtb & Case & 0.685 & 0.696 & 0.696 & 0.7  \\ 
et-ewt & Person & 0.609 & 0.696 & 0.609 & -  \\ 
et-ewt & Number & 0.551 & 0.551 & 0.48 & -  \\ 
et-ewt & Tense & 0.409 & 0.682 & 0.636 & -  \\ 
et-ewt & Mood & 0.533 & 0.4 & 0.533 & -  \\ 
et-ewt & Case & 0.7 & 0.754 & 0.657 & -  \\ 
fro-srcmf & Tense & 0.5 & 1.0 & 0.5 & 1.0  \\ 
es-gsd & Gender & 0.718 & 0.718 & 0.366 & 0.736  \\ 
es-gsd & Person & 0.591 & 0.545 & 0.591 & 0.355  \\ 
es-gsd & Number & 0.644 & 0.644 & 0.424 & 0.567  \\ 
es-gsd & Tense & 0.529 & 0.824 & 0.824 & 0.409  \\ 
es-gsd & Mood & 0.533 & 0.467 & 0.533 & 0.474  \\ 
es-gsd & Case & 0.0 & 1.0 & 0.0 & 0.0  \\

\end{tabular}
 }
 \caption{Comparing the ARM scores for SUD treebanks across both Statistical and Hard thresholding.}
 \label{tab:all_3}
 \end{center}

 \end{table*}

\begin{table*}[h]
\small
 \begin{center}{
 \begin{tabular}{l|l|l|l|l|l}
 \textbf{\textsc{treebank}} &  \textbf{\textsc{feature}} & \textbf{\textsc{Statistical}} & \textbf{\textsc{Hard}} &
 \textbf{\textsc{Baseline}} &
\textbf{\textsc{Dev}} \\
 \toprule
sl-ssj & Gender & 0.818 & 0.8 & 0.527 & 0.772  \\ 
sl-ssj & Person & 0.667 & 0.722 & 0.722 & 0.706  \\ 
sl-ssj & Number & 0.683 & 0.683 & 0.564 & 0.712  \\ 
sl-ssj & Tense & 0.333 & 0.583 & 0.333 & 0.364  \\ 
sl-ssj & Mood & 0.5 & 0.75 & 0.5 & 0.667  \\ 
sl-ssj & Case & 0.607 & 0.721 & 0.557 & 0.61  \\ 
cs-pdt & Gender & 0.564 & 0.788 & 0.75 & 0.58  \\ 
cs-pdt & Person & 0.591 & 0.705 & 0.614 & 0.541  \\ 
cs-pdt & Number & 0.477 & 0.64 & 0.629 & 0.481  \\ 
cs-pdt & Tense & 0.667 & 0.786 & 0.786 & 0.658  \\ 
cs-pdt & Mood & 0.538 & 0.538 & 0.538 & 0.48  \\ 
cs-pdt & Case & 0.646 & 0.675 & 0.545 & 0.633  \\ 
hsb-ufal & Gender & 0.857 & 0.714 & 0.786 & -  \\ 
hsb-ufal & Number & 0.692 & 0.538 & 0.692 & -  \\ 
hsb-ufal & Tense & 0.667 & 0.667 & 0.667 & -  \\ 
hsb-ufal & Case & 1.0 & 0.846 & 0.462 & -  \\ 
ga-idt & Gender & 0.64 & 0.76 & 0.78 & 0.647  \\ 
ga-idt & Person & 0.625 & 0.875 & 0.5 & 1.0  \\ 
ga-idt & Number & 0.468 & 0.571 & 0.468 & 0.446  \\ 
ga-idt & Tense & 0.714 & 0.571 & 0.429 & 0.5  \\ 
ga-idt & Mood & 0.833 & 0.833 & 0.667 & 0.714  \\ 
ga-idt & Case & 0.69 & 0.724 & 0.724 & 0.667  \\ 
gl-treegal & Gender & 0.722 & 0.685 & 0.333 & -  \\ 
gl-treegal & Person & 0.522 & 0.565 & 0.522 & -  \\ 
gl-treegal & Number & 0.68 & 0.546 & 0.361 & -  \\ 
gl-treegal & Tense & 0.462 & 0.538 & 0.692 & -  \\ 
gl-treegal & Mood & 0.462 & 0.692 & 0.462 & -  \\ 
fa-seraji & Person & 0.667 & 0.667 & 0.381 & 0.842  \\ 
fa-seraji & Number & 0.514 & 0.514 & 0.514 & 0.556  \\ 
fa-seraji & Tense & 0.455 & 0.545 & 0.636 & 0.545  \\ 
fa-seraji & Mood & 0.333 & 0.667 & 0.333 & 0.0  \\ 
et-edt & Person & 0.613 & 0.613 & 0.71 & 0.714  \\ 
et-edt & Number & 0.648 & 0.644 & 0.539 & 0.676  \\ 
et-edt & Tense & 0.579 & 0.632 & 0.763 & 0.537  \\ 
et-edt & Mood & 0.524 & 0.571 & 0.571 & 0.667  \\ 
et-edt & Case & 0.565 & 0.756 & 0.786 & 0.614  \\ 
la-perseus & Gender & 0.692 & 0.585 & 0.538 & -  \\ 
la-perseus & Person & 0.5 & 0.667 & 0.833 & -  \\ 
la-perseus & Number & 0.544 & 0.662 & 0.603 & -  \\ 
la-perseus & Tense & 0.75 & 1.0 & 1.0 & -  \\ 
la-perseus & Mood & 0.667 & 0.667 & 0.833 & -  \\ 
la-perseus & Case & 0.717 & 0.66 & 0.528 & -  \\ 
ug-udt & Person & 0.526 & 0.526 & 0.579 & 0.611  \\ 
ug-udt & Number & 0.767 & 0.6 & 0.533 & 0.697  \\ 
ug-udt & Tense & 0.625 & 0.75 & 0.5 & 0.778  \\ 
ug-udt & Mood & 0.692 & 0.923 & 0.769 & 0.833  \\ 
ug-udt & Case & 0.683 & 0.683 & 0.683 & 0.671  \\ 
es-ancora & Gender & 0.754 & 0.754 & 0.431 & 0.759  \\ 
es-ancora & Person & 0.429 & 0.429 & 0.429 & 0.526  \\ 
es-ancora & Number & 0.664 & 0.664 & 0.539 & 0.651  \\ 
es-ancora & Tense & 0.625 & 0.833 & 0.833 & 0.63  \\ 
es-ancora & Mood & 0.652 & 0.348 & 0.652 & 0.5  \\ 
de-hdt & Gender & 0.541 & 0.607 & 0.607 & 0.603  \\ 
de-hdt & Person & 0.071 & 0.929 & 0.071 & 0.085  \\ 
de-hdt & Number & 0.561 & 0.595 & 0.59 & 0.533  \\ 
de-hdt & Tense & 0.8 & 0.88 & 0.88 & 0.692  \\ 
de-hdt & Mood & 0.0 & 1.0 & 0.0 & 0.077  \\ 
de-hdt & Case & 0.738 & 0.836 & 0.574 & 0.7  \\ 
kk-ktb & Person & 0.636 & 0.545 & 0.636 & -  \\ 
kk-ktb & Number & 0.538 & 0.615 & 0.538 & -  \\ 
kk-ktb & Mood & 1.0 & 1.0 & 0.6 & -  \\ 
de-gsd & Gender & 0.699 & 0.781 & 0.397 & 0.641  \\ 
de-gsd & Person & 0.567 & 0.433 & 0.567 & 0.667  \\ 
de-gsd & Number & 0.638 & 0.638 & 0.35 & 0.619  \\ 
de-gsd & Tense & 0.455 & 0.636 & 0.591 & 0.526  \\ 
de-gsd & Mood & 0.5 & 0.455 & 0.455 & 0.421  \\ 
de-gsd & Case & 0.55 & 0.588 & 0.362 & 0.603  \\ 
nl-alpino & Gender & 0.667 & 0.8 & 0.8 & 0.562  \\ 
nl-alpino & Number & 0.548 & 0.548 & 0.565 & 0.625  \\ 
nl-alpino & Tense & 0.562 & 0.5 & 0.375 & 0.529  \\ 
af-afribooms & Number & 0.6 & 0.667 & 0.533 & 0.667  \\ 
af-afribooms & Tense & 0.842 & 0.842 & 0.842 & 0.588  \\ 
af-afribooms & Case & 0.0 & 1.0 & 0.0 & 0.0  \\

\end{tabular}
 }
 \caption{ Comparing the ARM scores for SUD treebanks across both Statistical and Hard thresholding.}
 \label{tab:all_4}
 \end{center}

 \end{table*}
 
\begin{table*}[h]
\small
 \begin{center}{
 \begin{tabular}{l|l|l|l|l|l}
 \textbf{\textsc{treebank}} &  \textbf{\textsc{feature}} & \textbf{\textsc{Statistical}} & \textbf{\textsc{Hard}} &
 \textbf{\textsc{Baseline}} &
\textbf{\textsc{Dev}} \\
 \toprule
 uk-iu & Gender & 0.701 & 0.693 & 0.559 & 0.771  \\ 
uk-iu & Person & 0.7 & 0.5 & 0.35 & 0.9  \\ 
uk-iu & Number & 0.647 & 0.659 & 0.479 & 0.656  \\ 
uk-iu & Tense & 0.476 & 0.476 & 0.571 & 0.615  \\ 
uk-iu & Mood & 0.318 & 0.409 & 0.318 & 0.357  \\ 
uk-iu & Case & 0.741 & 0.741 & 0.504 & 0.732  \\ 
cs-cltt & Gender & 0.857 & 0.929 & 0.75 & 0.806  \\ 
cs-cltt & Number & 0.646 & 0.688 & 0.479 & 0.576  \\ 
cs-cltt & Tense & 0.167 & 0.5 & 0.167 & 0.143  \\ 
cs-cltt & Mood & 0.0 & 1.0 & 0.0 & 0.0  \\ 
cs-cltt & Case & 0.697 & 0.758 & 0.636 & 0.658  \\ 
cop-scriptorium & Gender & 0.714 & 0.857 & 0.143 & 0.8  \\ 
cop-scriptorium & Number & 0.4 & 0.6 & 0.2 & 0.714  \\ 
ru-taiga & Gender & 0.648 & 0.724 & 0.638 & 0.667  \\ 
ru-taiga & Person & 0.667 & 0.75 & 0.583 & 0.786  \\ 
ru-taiga & Number & 0.662 & 0.601 & 0.459 & 0.646  \\ 
ru-taiga & Tense & 0.538 & 0.615 & 0.615 & 0.583  \\ 
ru-taiga & Mood & 0.611 & 0.667 & 0.611 & 0.5  \\ 
ru-taiga & Case & 0.557 & 0.696 & 0.633 & 0.593  \\ 
hu-szeged & Person & 0.444 & 0.556 & 0.444 & 0.138  \\ 
hu-szeged & Number & 0.396 & 0.64 & 0.396 & 0.434  \\ 
hu-szeged & Tense & 0.6 & 0.8 & 0.8 & 0.769  \\ 
hu-szeged & Mood & 0.714 & 0.714 & 0.714 & 0.5  \\ 
sr-set & Gender & 0.803 & 0.817 & 0.479 & 0.622  \\ 
sr-set & Person & 0.35 & 0.75 & 0.35 & 0.4  \\ 
sr-set & Number & 0.64 & 0.64 & 0.509 & 0.615  \\ 
sr-set & Tense & 0.474 & 0.684 & 0.684 & 0.444  \\ 
sr-set & Mood & 0.286 & 0.714 & 0.286 & 0.2  \\ 
sr-set & Case & 0.704 & 0.765 & 0.531 & 0.651  \\ 
en-lines & Person & 0.625 & 0.688 & 0.562 & 0.789  \\ 
en-lines & Number & 0.319 & 0.783 & 0.319 & 0.325  \\ 
en-lines & Tense & 0.704 & 0.778 & 0.704 & 0.636  \\ 
en-lines & Mood & 0.211 & 0.789 & 0.211 & 0.207  \\ 
en-lines & Case & 0.778 & 0.778 & 0.444 & 0.833  \\ 
sk-snk & Gender & 0.692 & 0.776 & 0.533 & 0.638  \\ 
sk-snk & Person & 0.778 & 0.333 & 0.222 & 0.625  \\ 
sk-snk & Number & 0.558 & 0.558 & 0.5 & 0.571  \\ 
sk-snk & Tense & 0.667 & 0.556 & 0.444 & 0.8  \\ 
sk-snk & Mood & 1.0 & 1.0 & 0.25 & 0.857  \\ 
sk-snk & Case & 0.731 & 0.756 & 0.526 & 0.833  \\ 
pl-pdb & Gender & 0.645 & 0.779 & 0.529 & 0.661  \\ 
pl-pdb & Person & 0.556 & 0.778 & 0.704 & 0.72  \\ 
pl-pdb & Number & 0.637 & 0.613 & 0.481 & 0.644  \\ 
pl-pdb & Tense & 0.5 & 0.6 & 0.7 & 0.6  \\ 
pl-pdb & Mood & 0.25 & 0.75 & 0.25 & 0.05  \\ 
pl-pdb & Case & 0.72 & 0.748 & 0.514 & 0.679  \\ 
la-ittb & Gender & 0.735 & 0.725 & 0.48 & 0.805  \\ 
la-ittb & Person & 0.19 & 0.81 & 0.19 & 0.273  \\ 
la-ittb & Number & 0.579 & 0.579 & 0.386 & 0.562  \\ 
la-ittb & Tense & 0.5 & 0.6 & 0.6 & 0.414  \\ 
la-ittb & Mood & 0.476 & 0.476 & 0.571 & 0.591  \\ 
la-ittb & Case & 0.757 & 0.796 & 0.495 & 0.792  \\ 
\end{tabular}
 }
 \caption{ Comparing the ARM scores for SUD treebanks across both Statistical and Hard thresholding.}
 \label{tab:all_6}
 \end{center}

 \end{table*}

\begin{table*}[h]
\small
 \begin{center}{
 \begin{tabular}{l|l|l|l|l|l}
 \textbf{\textsc{treebank}} &  \textbf{\textsc{feature}} & \textbf{\textsc{Statistical}} & \textbf{\textsc{Hard}} &
 \textbf{\textsc{Baseline}} &
\textbf{\textsc{Dev}} \\
 \toprule
da-ddt & Gender & 0.818 & 0.818 & 0.364 & 0.889  \\ 
da-ddt & Number & 0.667 & 0.667 & 0.286 & 0.725  \\ 
da-ddt & Tense & 0.737 & 0.737 & 0.842 & 0.562  \\ 
da-ddt & Mood & 0.2 & 0.8 & 0.2 & 0.077  \\ 
it-postwita & Gender & 0.702 & 0.702 & 0.362 & 0.674  \\ 
it-postwita & Person & 0.595 & 0.676 & 0.73 & 0.595  \\ 
it-postwita & Number & 0.744 & 0.744 & 0.558 & 0.642  \\ 
it-postwita & Tense & 0.481 & 0.704 & 0.704 & 0.607  \\ 
it-postwita & Mood & 0.792 & 0.792 & 0.792 & 0.556  \\ 
eu-bdt & Number & 0.508 & 0.6 & 0.415 & 0.473  \\ 
eu-bdt & Mood & 0.421 & 0.737 & 0.421 & 0.529  \\ 
eu-bdt & Case & 0.795 & 0.803 & 0.726 & 0.776  \\ 
sl-sst & Gender & 0.724 & 0.618 & 0.513 & -  \\ 
sl-sst & Person & 0.688 & 0.812 & 0.719 & -  \\ 
sl-sst & Number & 0.678 & 0.672 & 0.483 & -  \\ 
sl-sst & Tense & 0.48 & 0.76 & 0.48 & -  \\ 
sl-sst & Mood & 0.56 & 0.6 & 0.56 & -  \\ 
sl-sst & Case & 0.615 & 0.637 & 0.549 & -  \\ 
be-hse & Gender & 0.596 & 0.553 & 0.404 & 0.692  \\ 
be-hse & Person & 0.5 & 0.5 & 0.0 & 0.75  \\ 
be-hse & Number & 0.646 & 0.646 & 0.431 & 0.596  \\ 
be-hse & Tense & 0.429 & 0.429 & 0.571 & 0.333  \\ 
be-hse & Mood & 0.286 & 0.286 & 0.286 & 0.2  \\ 
be-hse & Case & 0.725 & 0.55 & 0.45 & 0.733  \\ 
fr-sequoia & Gender & 0.8 & 0.771 & 0.371 & 0.647  \\ 
fr-sequoia & Person & 0.667 & 0.667 & 0.4 & 0.857  \\ 
fr-sequoia & Number & 0.56 & 0.62 & 0.45 & 0.68  \\ 
fr-sequoia & Tense & 0.529 & 0.765 & 0.765 & 0.684  \\ 
fr-sequoia & Mood & 0.286 & 0.714 & 0.286 & 0.077  \\ 
sme-giella & Number & 0.653 & 0.653 & 0.561 & -  \\ 
sme-giella & Tense & 0.455 & 0.545 & 0.455 & -  \\ 
sme-giella & Mood & 0.214 & 0.571 & 0.214 & -  \\ 
sme-giella & Case & 0.741 & 0.704 & 0.667 & -  \\ 
el-gdt & Gender & 0.638 & 0.745 & 0.447 & 0.744  \\ 
el-gdt & Person & 0.667 & 0.667 & 0.458 & 0.667  \\ 
el-gdt & Number & 0.627 & 0.7 & 0.427 & 0.615  \\ 
el-gdt & Tense & 0.6 & 1.0 & 1.0 & 0.462  \\ 
el-gdt & Mood & 0.0 & 1.0 & 0.0 & 0.0  \\ 
el-gdt & Case & 0.809 & 0.809 & 0.319 & 0.814  \\ 
orv-torot & Gender & 0.655 & 0.669 & 0.547 & 0.679  \\ 
orv-torot & Person & 0.6 & 0.6 & 0.6 & 0.594  \\ 
orv-torot & Number & 0.621 & 0.621 & 0.581 & 0.618  \\ 
orv-torot & Tense & 0.731 & 0.731 & 0.769 & 0.72  \\ 
orv-torot & Mood & 0.316 & 0.789 & 0.421 & 0.176  \\ 
orv-torot & Case & 0.709 & 0.775 & 0.609 & 0.691  \\ 
sv-lines & Gender & 0.538 & 0.538 & 0.308 & 0.64  \\ 
sv-lines & Number & 0.643 & 0.643 & 0.452 & 0.529  \\ 
sv-lines & Tense & 0.429 & 0.476 & 0.429 & 0.655  \\ 
sv-lines & Mood & 0.231 & 0.769 & 0.231 & 0.161  \\ 
sv-lines & Case & 0.583 & 0.583 & 0.25 & 0.51  \\ 
ta-ttb & Gender & 0.682 & 0.682 & 0.659 & 0.5  \\ 
ta-ttb & Person & 0.091 & 0.955 & 0.091 & 0.167  \\ 
ta-ttb & Number & 0.523 & 0.591 & 0.545 & 0.533  \\ 
ta-ttb & Tense & 0.625 & 0.5 & 0.625 & 0.667  \\ 
ta-ttb & Mood & 0.5 & 1.0 & 0.5 & 0.5  \\ 
ta-ttb & Case & 0.846 & 0.846 & 0.692 & 1.0  \\ 
it-partut & Gender & 0.786 & 0.786 & 0.25 & 0.846  \\ 
it-partut & Person & 0.833 & 0.917 & 0.25 & 0.615  \\ 
it-partut & Number & 0.714 & 0.508 & 0.286 & 0.576  \\ 
it-partut & Tense & 0.9 & 0.9 & 0.6 & 0.583  \\ 
it-partut & Mood & 0.2 & 0.4 & 0.2 & 0.167  \\ 
ar-padt & Gender & 0.592 & 0.592 & 0.549 & 0.712  \\ 
ar-padt & Person & 0.0 & 0.833 & 0.0 & 0.263  \\ 
ar-padt & Number & 0.512 & 0.643 & 0.512 & 0.593  \\ 
ar-padt & Mood & 0.571 & 0.571 & 0.571 & 0.6  \\ 
ar-padt & Case & 0.871 & 0.871 & 0.753 & 0.824  \\ 
bg-btb & Gender & 0.638 & 0.66 & 0.404 & 0.585  \\ 
bg-btb & Person & 0.625 & 0.625 & 0.625 & 0.625  \\ 
bg-btb & Number & 0.639 & 0.631 & 0.533 & 0.679  \\ 
bg-btb & Tense & 0.6 & 0.6 & 0.6 & 0.579  \\ 
bg-btb & Mood & 0.056 & 0.944 & 0.056 & 0.176  \\ 
 \end{tabular}
 }
 \caption{ Comparing the ARM scores for SUD treebanks across both Statistical and Hard thresholding.}
 \label{tab:all_7}
 \end{center}

 \end{table*}
 
\begin{table*}[h]
\small
 \begin{center}{
 \begin{tabular}{l|l|l|l|l|l}
 \textbf{\textsc{treebank}} &  \textbf{\textsc{feature}} & \textbf{\textsc{Statistical}} & \textbf{\textsc{Hard}} &
 \textbf{\textsc{Baseline}} &
\textbf{\textsc{Dev}} \\
 \toprule
 pt-bosque & Gender & 0.656 & 0.721 & 0.41 & 0.792  \\ 
pt-bosque & Person & 0.25 & 0.75 & 0.25 & 0.455  \\ 
pt-bosque & Number & 0.669 & 0.669 & 0.378 & 0.698  \\ 
pt-bosque & Tense & 0.5 & 0.438 & 0.438 & 0.692  \\ 
pt-bosque & Mood & 0.375 & 0.5 & 0.375 & 0.429  \\ 
lt-alksnis & Gender & 0.711 & 0.711 & - & 0.671  \\ 
lt-alksnis & Person & 0.667 & 0.8 & 0.667 & 0.571  \\ 
lt-alksnis & Number & 0.625 & 0.625 & 0.531 & 0.595  \\ 
lt-alksnis & Tense & 0.667 & 0.667 & 0.667 & 0.6  \\ 
lt-alksnis & Mood & 0.667 & 0.333 & 0.667 & 0.375  \\ 
lt-alksnis & Case & 0.826 & 0.826 & 0.496 & 0.798  \\ 
ar-nyuad & Gender & 0.606 & 0.718 & 0.718 & 0.536  \\ 
ar-nyuad & Person & 0.469 & 0.562 & 0.469 & 0.343  \\ 
ar-nyuad & Number & 0.502 & 0.554 & 0.502 & 0.468  \\ 
ar-nyuad & Mood & 0.438 & 0.562 & 0.438 & 0.5  \\ 
ar-nyuad & Case & 0.627 & 0.747 & 0.747 & 0.551  \\ 
ca-ancora & Gender & 0.804 & 0.786 & 0.464 & 0.77  \\ 
ca-ancora & Person & 0.389 & 0.611 & 0.389 & 0.219  \\ 
ca-ancora & Number & 0.652 & 0.652 & 0.511 & 0.616  \\ 
ca-ancora & Tense & 0.5 & 0.731 & 0.692 & 0.56  \\ 
ca-ancora & Mood & 0.32 & 0.68 & 0.32 & 0.348  \\ 
grc-proiel & Gender & 0.605 & 0.516 & 0.535 & 0.588  \\ 
grc-proiel & Person & 0.543 & 0.543 & 0.6 & 0.737  \\ 
grc-proiel & Number & 0.533 & 0.61 & 0.538 & 0.585  \\ 
grc-proiel & Tense & 0.643 & 0.786 & 0.786 & 0.774  \\ 
grc-proiel & Mood & 0.529 & 0.529 & 0.529 & 0.65  \\ 
grc-proiel & Case & 0.809 & 0.854 & 0.51 & 0.813  \\ 
it-twittiro & Gender & 0.808 & 0.808 & 0.385 & 0.65  \\ 
it-twittiro & Person & 0.591 & 0.318 & 0.682 & 0.579  \\ 
it-twittiro & Number & 0.568 & 0.568 & 0.419 & 0.634  \\ 
it-twittiro & Tense & 0.25 & 0.75 & 0.75 & 0.462  \\ 
it-twittiro & Mood & 0.5 & 0.5 & 0.5 & 0.364  \\ 
mr-ufal & Gender & 0.609 & 0.652 & 0.565 & 0.52  \\ 
mr-ufal & Person & 0.727 & 0.727 & 0.364 & 0.889  \\ 
mr-ufal & Number & 0.394 & 0.794 & 0.242 & 0.514  \\ 
mr-ufal & Case & 0.583 & 0.583 & 0.417 & 0.857  \\ 
tr-imst & Person & 0.359 & 0.818 & 0.359 & 0.342  \\ 
tr-imst & Number & 0.47 & 0.536 & 0.47 & 0.485  \\ 
tr-imst & Tense & 0.762 & 0.762 & 0.81 & 0.68  \\ 
tr-imst & Mood & 0.714 & 0.714 & 0.714 & 0.68  \\ 
tr-imst & Case & 0.717 & 0.804 & 0.804 & 0.678  \\ 
bxr-bdt & Case & 0.818 & 0.545 & 0.818 & -  \\ 
hi-hdtb & Gender & 0.586 & 0.617 & 0.5 & 0.631  \\ 
hi-hdtb & Person & 0.045 & 0.955 & 0.045 & 0.052  \\ 
hi-hdtb & Number & 0.416 & 0.615 & 0.416 & 0.455  \\ 
hi-hdtb & Tense & 0.333 & 0.333 & 0.333 & 0.2  \\ 
hi-hdtb & Mood & 1.0 & 1.0 & 0.333 & 0.667  \\ 
hi-hdtb & Case & 0.654 & 0.709 & 0.63 & 0.62  \\ 
hr-set & Gender & 0.725 & 0.717 & 0.525 & 0.643  \\ 
hr-set & Person & 0.769 & 0.769 & 0.577 & 0.692  \\ 
hr-set & Number & 0.675 & 0.675 & 0.51 & 0.658  \\ 
hr-set & Tense & 0.429 & 0.714 & 0.714 & 0.542  \\ 
hr-set & Mood & 0.412 & 0.588 & 0.412 & 0.158  \\ 
hr-set & Case & 0.669 & 0.725 & 0.577 & 0.659  \\ 
kmr-mg & Gender & 1.0 & 0.818 & 1.0 & -  \\ 
kmr-mg & Number & 0.783 & 0.739 & 0.783 & -  \\ 
kmr-mg & Case & 0.909 & 0.727 & 0.909 & -  \\ 
nl-lassysmall & Gender & 0.85 & 0.85 & 0.9 & 0.81  \\ 
nl-lassysmall & Number & 0.646 & 0.646 & 0.523 & 0.646  \\ 
nl-lassysmall & Tense & 0.6 & 0.6 & 0.4 & 0.364  \\ 
fr-gsd & Gender & 0.727 & 0.727 & 0.485 & 0.807  \\ 
fr-gsd & Person & 0.375 & 0.719 & 0.375 & 0.312  \\ 
fr-gsd & Number & 0.624 & 0.624 & 0.441 & 0.593  \\ 
fr-gsd & Tense & 0.706 & 0.706 & 0.765 & 0.81  \\ 
fr-gsd & Mood & 0.273 & 0.727 & 0.273 & 0.25  \\ 
 
 \end{tabular}
 }
 \caption{ Comparing the ARM scores for SUD treebanks across both Statistical and Hard thresholding.}
 \label{tab:all_8}
 \end{center}

 \end{table*}
 
\begin{table*}[h]
\small
 \begin{center}{
 \begin{tabular}{l|l|l|l|l|l}
 \textbf{\textsc{treebank}} &  \textbf{\textsc{feature}} & \textbf{\textsc{Statistical}} & \textbf{\textsc{Hard}} &
 \textbf{\textsc{Baseline}} &
\textbf{\textsc{Dev}} \\
 got-proiel & Gender & 0.559 & 0.595 & 0.559 & 0.658  \\ 
got-proiel & Person & 0.571 & 0.771 & 0.657 & 0.614  \\ 
got-proiel & Number & 0.64 & 0.68 & 0.503 & 0.591  \\ 
got-proiel & Tense & 0.714 & 0.714 & 0.714 & 0.586  \\ 
got-proiel & Mood & 0.722 & 0.722 & 0.611 & 0.682  \\ 
got-proiel & Case & 0.82 & 0.784 & 0.505 & 0.803  \\ 
en-gum & Person & 0.167 & 0.917 & 0.167 & 0.176  \\ 
en-gum & Number & 0.397 & 0.767 & 0.397 & 0.259  \\ 
en-gum & Tense & 0.579 & 0.684 & 0.579 & 0.625  \\ 
en-gum & Mood & 0.176 & 0.824 & 0.176 & 0.05  \\ 
lzh-kyoto & Mood & 0.0 & 1.0 & 0.0 & 0.0  \\ 
lzh-kyoto & Case & 0.0 & 1.0 & 0.0 & 0.125  \\ 
cs-fictree & Gender & 0.717 & 0.683 & 0.4 & 0.691  \\ 
cs-fictree & Person & 0.667 & 0.905 & 0.81 & 0.625  \\ 
cs-fictree & Number & 0.649 & 0.649 & 0.364 & 0.673  \\ 
cs-fictree & Tense & 0.833 & 0.889 & 0.778 & 0.565  \\ 
cs-fictree & Mood & 0.455 & 0.455 & 0.545 & 0.643  \\ 
cs-fictree & Case & 0.697 & 0.652 & 0.461 & 0.738  \\ 
hy-armtdp & Person & 0.444 & 0.593 & 0.593 & 0.692  \\ 
hy-armtdp & Number & 0.592 & 0.612 & 0.561 & 0.676  \\ 
hy-armtdp & Tense & 0.824 & 0.765 & 0.529 & 0.733  \\ 
hy-armtdp & Mood & 0.789 & 0.789 & 0.737 & 0.8  \\ 
hy-armtdp & Case & 0.857 & 0.857 & 0.821 & 0.772  \\ 
gd-arcosg & Gender & 0.615 & 0.615 & 0.615 & 0.609  \\ 
gd-arcosg & Person & 0.6 & 0.8 & 0.6 & 0.75  \\ 
gd-arcosg & Number & 0.562 & 0.562 & 0.562 & 0.588  \\ 
gd-arcosg & Tense & 0.833 & 0.333 & 0.5 & 0.8  \\ 
gd-arcosg & Mood & 0.667 & 0.667 & 0.333 & 0.714  \\ 
gd-arcosg & Case & 0.85 & 0.85 & 0.5 & 0.833  \\ 
lt-hse & Gender & 0.658 & 0.553 & 0.474 & 0.6  \\ 
lt-hse & Person & 0.778 & 0.444 & 0.444 & 0.8  \\ 
lt-hse & Number & 0.642 & 0.597 & 0.478 & 0.667  \\ 
lt-hse & Tense & 0.714 & 0.857 & 0.857 & 0.889  \\ 
lt-hse & Mood & 0.2 & 0.6 & 0.2 & 0.429  \\ 
lt-hse & Case & 0.564 & 0.615 & 0.641 & 0.816  \\ 
no-nynorsklia & Gender & 0.727 & 0.697 & 0.455 & 0.667  \\ 
no-nynorsklia & Person & 1.0 & 1.0 & 0.0 & 1.0  \\ 
no-nynorsklia & Number & 0.743 & 0.743 & 0.343 & 0.649  \\ 
no-nynorsklia & Tense & 0.435 & 0.826 & 0.783 & 0.435  \\ 
no-nynorsklia & Mood & 0.0 & 1.0 & 0.0 & 0.043  \\ 
no-nynorsklia & Case & 0.5 & 1.0 & 0.5 & 0.0  \\ 
cu-proiel & Gender & 0.61 & 0.66 & 0.54 & 0.706  \\ 
cu-proiel & Person & 0.667 & 0.667 & 0.528 & 0.579  \\ 
cu-proiel & Number & 0.672 & 0.579 & 0.503 & 0.641  \\ 
cu-proiel & Tense & 0.567 & 0.533 & 0.6 & 0.655  \\ 
cu-proiel & Mood & 0.348 & 0.652 & 0.348 & 0.364  \\ 
cu-proiel & Case & 0.818 & 0.818 & 0.473 & 0.793  \\

  \end{tabular}
 }
 \caption{ Comparing the ARM scores for SUD treebanks across both Statistical and Hard thresholding.}
 \label{tab:all_9}
 \end{center}

 \end{table*}

\end{document}